%% file: preprint.tex
\documentclass{article}
\usepackage{iclr2027_conference,times}

\input{math_commands.tex}

\usepackage[hidelinks,colorlinks=false]{hyperref}
\usepackage{url}
\usepackage{graphicx}
\usepackage{amsmath}
\usepackage{amssymb}
\usepackage{algorithm}
\usepackage{algorithmic}
\usepackage{adjustbox}
\usepackage{booktabs}
\usepackage{enumitem}
\definecolor{linkblue}{HTML}{355C7D}\hypersetup{colorlinks=true,linkcolor=linkblue,citecolor=linkblue,urlcolor=linkblue}
\usepackage{wrapfig}
\usepackage[table]{xcolor}
\title{Not All Errors Are Equal: Consequence-Aware Reasoning Compute Allocation}
\author{
  Liang He,\hspace{0.4em}
  Jingbo Wen,\hspace{0.4em}
  Haoyu Wang,\hspace{0.4em}
  Ziqi He,\hspace{0.4em}
  Yixiong Chen,\hspace{0.4em}
  Kangning Cui,\hspace{0.4em}
  Xilu Wang
  \\[0.8em]
  {\small
    University of Sydney \hspace{0.4em}
    Johns Hopkins University \hspace{0.4em}
    City University of Hong Kong\hspace{0.4em}
    University of Surrey
  }
}

\iclrfinalcopy

\begin{document}
\maketitle

\lhead{}
\renewcommand{\headrulewidth}{0pt}
\begin{abstract}
Test-time compute has emerged as an effective paradigm for improving large language model capability at inference time. Existing allocation strategies primarily prioritize tasks according to difficulty,
uncertainty, or expected performance gain, implicitly treating prediction errors
as equally costly. This assumption is often misaligned with real deployment, where failures can differ substantially in their downstream tasks. To address this limitation, this paper introduces \emph{consequence-aware test-time compute allocation} by formulating a cost-weighted scheduling problem where the priority of a task is its failure consequence with the marginal gain of additional compute. In practice, however, marginal gain is difficult to predict before execution, so we
propose a deployable scheduler that uses consequence as the routing signal. The scheduler predicts task consequence from pre-solution inputs and allocates the available premium compute to the corresponding top-ranked tasks.  Experiments on the SWE bench Lite show that consequence provides information beyond task difficulty and can be predicted before solving. Under a fixed compute budget, consequence-aware routing achieves the best high-consequence task success, while overall accuracy remains competitive. A controlled within-model experiment further confirms the same advantage when only inference attempts are
reallocated.  

\end{abstract}
\section{Introduction}
\label{sec:intro}

Recent advances in large language models (LLMs) have established test-time computation as an important scaling dimension beyond conventional training-time scaling through model capacity and data \citep{geiping2026scaling,setlur2026e3}. Increasing inference-time computation can improve model performance through longer reasoning traces \citep{wei2022chain}, multiple sampling attempts \citep{wang2023selfconsistency}, process-reward verification \citep{lightman2024let}, or routing to stronger models \citep{ong2025routellm,zhang2026let}. Since applying these additional computation uniformly across all queries can be costly and unnecessary, a growing body of work studies adaptive test-time compute allocation and focuses on \emph{how to distribute a limited inference budget across tasks}.

Existing approaches largely answer this question using task difficulty \citep{snell2024scaling,damani2024learning}, model uncertainty or confidence \citep{manvi2024adaptive}, and predicted gains from additional computation \citep{qu2026adaptivetesttime}.
Tasks that appear harder, less certain, or more likely to benefit from
additional computation receive more resources. These signals focus on the likelihood of failure or the expected benefit of additional compute, but not on the consequence of failing a particular task. Optimizing the allocation for averaged accuracy across tasks is natural for benchmark evaluation, as each task contributes equally to the final score and all failures are treated the same. In deployment, however, task failures can incur different real-world consequences that average accuracy does not capture. Consider a coding assistant: it might introduce a documentation error or produce a faulty database migration. In a benchmark, both are counted as a single failure, but their real-world consequences differ substantially. The former may cause brief user confusion, whereas the latter can corrupt production data and require weeks of recovery. Allocating computation solely based on failure likelihood, without accounting for the \emph{consequences of failure}, can leave costly yet preventable failures insufficiently addressed, even when additional compute is available.

\begin{figure}[t]
\centering
\includegraphics[width=0.9\linewidth]{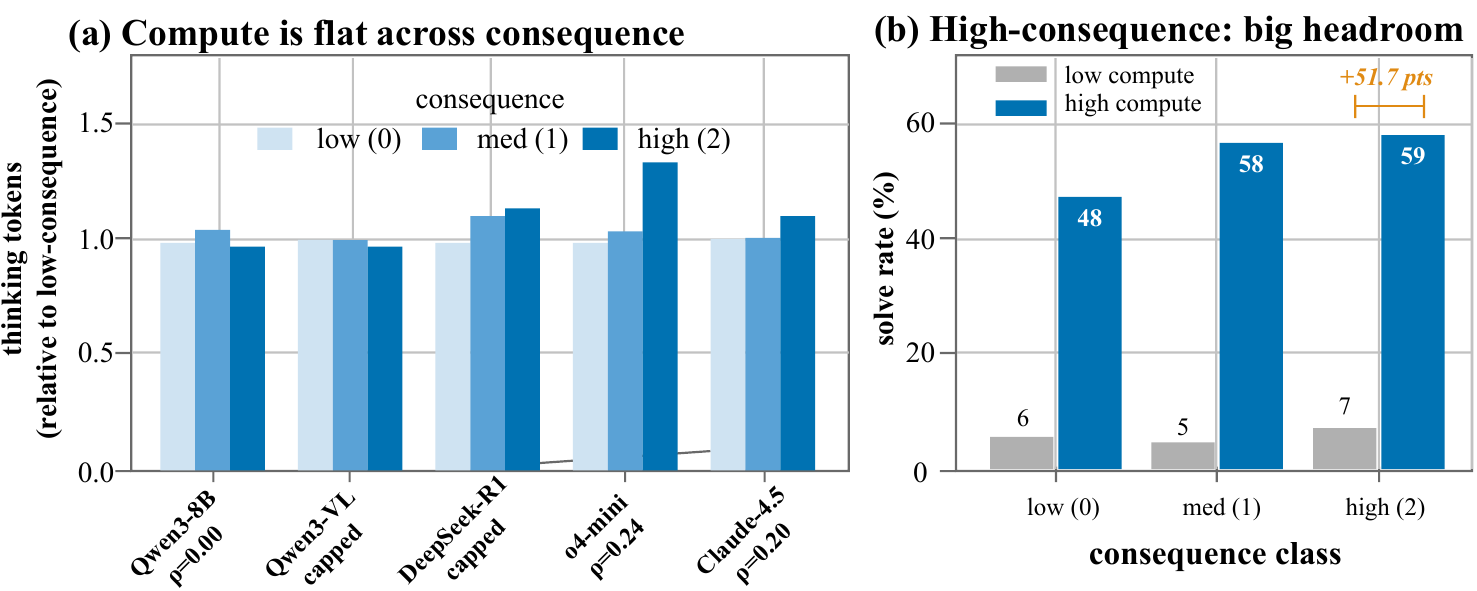}
\caption{\textbf{Native reasoning compute does not reliably track consequence despite available headroom.}
(a) Across five reasoning models, native computation is weakly aligned
with consequence or saturates the token budget. Numbers report Spearman
correlation or the fraction of tasks reaching the cap.
(b) On SWE-bench Lite, class-2 success increases from $7\%$ under the
cheap tier to $59\%$ under the premium tier in the 16-system setting.}
\label{fig:motivation}
\end{figure}

This motivates us to introduce a new signal, \emph{consequence of failure}, to guide adaptive test-time compute allocation. We define \emph{consequence of failure} as the deployment impact of an incorrect answer generated by LLMs, i.e., the cost incurred if the LLM produces a wrong solution and the system deploys it. Unlike difficulty or uncertainty, \emph{consequence of failure} depends on the task and its deployment context, rather than only on how likely the model is to be correct. We first ask whether current reasoning models already account for consequence in their native use of test-time compute. Across five contemporary reasoning models, we find little consistent relationship between native computation and task consequence: some models show weak or no correlation, while others frequently saturate their token budgets (Figure~\ref{fig:motivation}(a)). We then ask whether high-consequence tasks can actually benefit from additional compute. In the 16-system compute tier for SWE-bench Lite, class-2 success rises from $7\%$ under the cheap tier to $59\%$ under the premium tier (Figure~\ref{fig:motivation}(b)), showing substantial headroom. Finally, we examine whether consequence simply reflects task difficulty. It does not: across multiple labeling strategies, consequence is nearly orthogonal to difficulty, and high-consequence tasks span the full difficulty spectrum. This pattern is consistent across human annotations, alternative rubrics, maintainer-derived severity, and cost-model sensitivity analyses (Appendices~\ref{app:iaa}, \ref{app:rubric}, \ref{app:realcost}, and~\ref{app:costmap}).



Motivated by these observations, we introduce consequence-aware test-time compute allocation, which explicitly accounts for the cost of failure under a fixed compute budget. More specifically, we design a lightweight predictor to estimate consequences from deployment-time inputs, without access to gold patches or execution outcomes, and a scheduler to allocate higher-consequence tasks with more computing resources. Unlike difficulty or uncertainty, which indicate where a model is likely to fail, consequence captures how costly a failure would be and can be predicted from information available before execution, making it a practical deployment-time signal when marginal gains are difficult to estimate reliably (Appendices~\ref{app:marginal} and~\ref{app:whyfail}).
Our deployable scheduler uses consequence alone as the routing signal;
priority-aware variants that additionally use empirical marginal gain
are reported only as diagnostic upper bounds.
On SWE-bench Lite, consequence-aware routing shifts success toward high-consequence tasks without improving overall accuracy, and a deployment-time predictor retains most of the oracle gain. Additional experiments examine transfer, alternative compute, and boundary conditions (Appendices~\ref{app:crossdataset}, \ref{app:seconddomain}, \ref{app:within}, and~\ref{app:ood}).
The main contributions of this paper are as follows:
\begin{itemize}[leftmargin=*,itemsep=1pt,topsep=2pt]

\item  We introduce the \emph{consequence of failure} as a distinct deployment signal for adaptive test time compute allocation. Unlike existing signals
that characterize failure likelihood or expected improvement, consequence captures the value of avoiding an error and shifts allocation from average accuracy toward deployment cost.

\item We develop a cost weighted formulation of compute allocation and derive the corresponding optimal priority rule in the two tier setting. Based on real deployment, we propose a deployable consequence aware allocation algorithm that predicts consequence from pre-solution inputs and reallocates a fixed budget toward higher-consequence tasks

\item We empirically establish that consequence is both distinct from existing routing signals and available at deployment time. Across SWE bench Lite, consequence shows little association with task difficulty, is not reflected
reliably in native reasoning compute, and can be recovered from pre solution task information.


\item Extensive experiments on SWE-bench Lite show that consequence-aware routing reallocates success toward costly failures under a matched budget, raising high-consequence task success from $19.4\%$ under random routing to
$51.1\%$ with overall accuracy unchanged, and a controlled within-model experiment reproduces this result when only inference attempts are reallocated.

\end{itemize}

\section{Related Work}

\textbf{Adaptive test-time compute.}
Scaling test-time compute has emerged as an effective way to improve LLM performance by spending additional computation at inference time. Existing techniques include longer reasoning traces, repeated sampling such as Best-of-$N$, verification and reward-guided decoding, self-refinement, and search-based methods~\citep{snell2024scaling}. While these approaches typically increase the compute devoted to a single query, more recent work has moved from uniform scaling toward adaptive test-time compute allocation, where the amount of computation varies across inputs. Existing methods use signals such as predicted task difficulty~\citep{snell2024scaling,damani2024learning}, model uncertainty or confidence~\citep{manvi2024adaptive}, and predicted gains from additional computation~\citep{qu2026adaptivetesttime} to decide which queries should receive more resources. A related line of work routes queries across model cascades, using cheaper models for simpler inputs and escalating to stronger models or more expensive decoding procedures when predicted quality is insufficient~\citep{chen2023frugalgpt,aggarwal2024automix,ong2024routellm,ding2024hybrid}. These methods differ in how compute is increased or routed, but they are largely accuracy-driven. Our work studies a new signal, the consequence of failure, and redistributes compute across queries when some failures have substantially greater deployment impact than others.

\textbf{Rational metareasoning.}
The decision-theoretic view of computation as a costly action dates back to rational metareasoning, which studies when additional computation is worth its expected benefit. Consequence-aware
allocation is a direct instantiation of this idea for modern reasoning
models: the utility of extra computation is the reduction in
consequence-weighted error, not merely the reduction in average error. Recent work has connected this perspective to LLMs by introducing both correctness and cost  (usually time or energy) to a reward function. 
We instead make the task-dependent cost of
being wrong explicit. In multi-task deployment, the value of additional computation depends not only on whether it can improve the outcome and how much computation it costs, but also on how serious a failure on that particular task would be.

\textbf{Cost-sensitive learning and selective prediction.}
Cost-sensitive learning recognizes that different errors can have
different costs~\citep{elkan2001foundations}, while selective
prediction allows a model to abstain when uncertainty is high
\citep{geifman2017selective}. These settings modify the prediction
rule or decide whether to answer. Our setting is different: the model
still answers every task, but the system decides how much computation
to spend before answering. This makes consequence-aware allocation
compatible with existing reasoning models, since the intervention is a
scheduler rather than a retraining objective.

\section{Problem Formulation and Method}
\subsection{Consequence of Failure}
\label{sec:cons}

\textbf{Definition.}
For a task $x$, we define its \emph{consequence of failure} as the
deployment impact of an incorrect answer. We represent it using an
ordinal severity label $s_x\in\{0,1,2\}$, corresponding to low,
medium, and high consequence, respectively.
Low-consequence failures ($s_x=0$) include cosmetic, formatting, documentation, or log-message failures that may be visible but do not corrupt downstream systems. Medium-consequence failures ($s_x=1$) include contained functional
failures, such as wrong return values, edge-case crashes, or locally
recoverable feature errors. High-consequence failures ($s_x=2$) include
silent data corruption, security or permission bypass, irreversible
deletion, overwriting, migration errors, or incorrect results consumed
downstream without timely visibility. We use this setting because it applies consistently to both LLM judges and human annotators. Appendix~\ref{app:costmap} shows that the qualitative allocation results are stable under alternative monotone mappings and continuous severity scores.
For allocation, we map the ordinal label to a numerical consequence
weight $w_x=g(s_x)$. In the main experiments, we use the simple mapping
$g(s)=s$, yielding $w_x\in\{0,1,2\}$. Class~0 serves as a normalized
reference level rather than implying zero real-world harm.
Alternative monotone mappings are evaluated in Appendix~\ref{app:costmap}.

\textbf{Labeling pipelines.}
The consequence of task failure is not necessarily known before solving it. We use three labeling modes. First, an LLM-with-patch judge observes
the issue text, the gold patch, and the modified files, and assigns the
reference consequence label used for the main cost-weighted evaluation.
Second, an issue-only predictor observes only deployment-time
information: the GitHub issue text and file paths explicitly mentioned
in the issue report itself, such as stack traces, error logs, or
user-specified file references. It does not receive the gold patch, the
reference modified-file list, any file path inferred from the gold
diff, or any generated fix. Third, a simple rule-based labeler provides
a lexical baseline. Human annotations, alternative labelers, and
maintainer-revealed severity are used only as construct-validity checks
in the appendix (Appendices~\ref{app:iaa}, \ref{app:rubric},
and~\ref{app:realcost}).

\textbf{Software-engineering benchmarks.}
We use SWE-bench Lite~\citep{jimenez2023swebench} as the primary testbed because software tasks naturally vary both in difficulty and in deployment consequence. A formatting bug, a parser bug, a permission bug, and a database migration error may all appear as single pass/fail benchmark tasks, while their real-world costs differ
substantially. The benchmark also provides issue reports, file paths, patches, and public solver outcomes, which allow us to compare deployment-time consequence prediction with with-patch reference labels and to evaluate compute allocation under matched budgets. We validate the construct and routing results with human labels,
maintainer behavior, alternative judges, a second database/ORM software pool, and out-of-domain probes in the appendix (Appendices~\ref{app:iaa}--\ref{app:context}).

\subsection{Method: Cost-Weighted Compute Allocation}
\label{sec:method}
\textbf{Setup.}
Let $X$ be a stream of tasks and let $\{T_1,\ldots,T_K\}$ denote $K$ compute tiers ordered by compute cost. A tier can correspond to a stronger model, more sampling attempts, a verifier-backed procedure, or another deployer-controlled compute level. Let $\kappa_k$ denote the compute cost of tier $T_k$, and let $B$
denote the total compute budget.

\textbf{Objective.}
The following objective formalizes the role of consequence in compute
routing; it is not tied to any particular model, inference mechanism, or
compute tier. Standard adaptive-compute methods implicitly optimize
average accuracy under a budget. This corresponds to treating all errors
as equally costly. 
We instead minimize consequence-weighted error:
\begin{equation}
\label{eq:budgetobj}
\min_{T:\,X\to\{T_1,\dots,T_K\}}
\sum_{x\in X}
w_x
\left[
1-p\bigl(\mathrm{correct}\mid x,T(x)\bigr)
\right]
\quad
\mathrm{s.t.}
\quad
\sum_{x\in X}\kappa_{T(x)} \le B .
\end{equation}
Here $w_x=g(s_x)$ is the consequence weight and
$\kappa_{T(x)}$ is the compute cost of the assigned tier. If $w_x$ is
constant across tasks, the objective reduces to standard
average-error allocation.

\textbf{Exact priority rule.}
Consider two compute tiers, a cheap tier $T_1$ and a premium tier $T_2$. Let
$\Delta_x = p_{\mathrm{premium}}(x)-p_{\mathrm{cheap}}(x)$ denote the marginal
success gain from premium compute, so that upgrading task $x$ from $T_1$ to
$T_2$ reduces the consequence-weighted objective by exactly $w_x\Delta_x$ while
incurring the same additional cost for every task. The budget therefore fixes the number of upgrades rather than which tasks receive them, and the exact optimum is obtained by assigning premium compute to the top-ranked tasks with the largest positive $w_x\Delta_x$. Optimal allocation therefore
depends jointly on consequence and marginal compute benefit. The weight $w_x$ determines how valuable it is to avoid an error, whereas $\Delta_x$ determines how much premium compute can improve the outcome. 

\textbf{Why Difficulty-Aware Routing Fails}
The exact priority rule also helps explain why difficulty-aware routing can fail. Difficulty-aware routing fails when task hardness is misaligned with $\Delta_x$. We measure task difficulty on SWE-bench Lite as $1-\overline{\mathrm{passrate}}$, where $\overline{\mathrm{passrate}}$ is the mean success rate of the task across the 16 public SWE-bench systems. Appendix~\ref{app:whyfail} shows on SWE-bench Lite, difficulty is strongly negatively correlated with $\Delta_x$ ($\rho=-0.78$, $p<10^{-60}$): many hard tasks remain unsolved even by premium systems. This means that allocating extra compute to the hardest tasks is largely wasteful. By contrast, consequence is nearly uncorrelated with marginal gain ($\rho=+0.09$, n.s.). Appendix~\ref{app:whyfail} shows that the negative difficulty--gain relationship persists after removing never-solved tasks, while deployment-time difficulty estimates remain weakly related to marginal gain; Appendix~\ref{app:marginal} further shows that $\Delta_x$ cannot be predicted reliably before execution.


\textbf{Why the deployable scheduler uses consequence only.}
The exact optimum requires estimating the marginal benefit of additional computation $\Delta_x$. However, $\Delta_x$ is not available before solving a task. Appendix~\ref{app:marginal} shows that both prompted forecasts and learned embedding probes fail to reliably predict marginal gain from issue text alone. Hence, our deployable scheduler ranks tasks by the predicted weight $\hat{w}_x=g(\hat{s}_x)$ alone, with details in Appendix~\ref{app:scheduler}. In the fixed-quota experiments of \S\ref{sec:main}, the top-$q\%$ tasks ranked by
$\hat{w}_x$ receive premium computation, where $q$ is selected to match the total compute budget of difficulty-aware baselines. 
Because the scheduler operates entirely at the inference layer, it
requires no model retraining, fine-tuning, or architectural changes.
It only requires a ranking of tasks by expected error consequence,
rather than calibrated monetary costs. This design allows consequence
awareness to be incorporated into existing reasoning systems without
modifying the underlying model.


\section{Experiments}
We organize our experiments around three questions, namely
(i) whether consequence carries information beyond difficulty and native
reasoning and can be predicted before solving (\S\ref{sec:signal}),
(ii) whether routing by predicted consequence protects high-consequence tasks under a fixed budget, both across systems (\S\ref{sec:main}) and within a single fixed model (\S\ref{sec:singlemodel}), and (iii) how robust the gain is and when it fails (\S\ref{sec:boundary}).
The first question validates consequence as a routing signal, and the remaining ones test its effectiveness and limits.

\subsection{Empirical Validation of the Consequence Signal}
\label{sec:signal}

\begin{wraptable}{r}{0.5\linewidth}
\centering
\caption{\textbf{Consequence carries little information about task
difficulty.} Spearman correlations stay within $\pm 0.11$ across rule-based,
LLM-based, deployment-time, and human labeling pipelines.}
\label{tab:consequence_orthogonality}
\footnotesize
\setlength{\tabcolsep}{3pt}
\renewcommand{\arraystretch}{1.0}
\begin{tabular}{@{}lrrr@{}}
\toprule
Labeling pipeline & $n$ & Spearman $\rho$ & $p$-value \\
\midrule
\rowcolor{gray!12}
\multicolumn{4}{@{}c}{\textit{Automated labeling}} \\
Rule-based           & $300$ & $+0.047$ & $0.42$ \\
LLM-with-patch       & $300$ & $-0.096$ & $0.10$ \\
Issue-only predictor & $300$ & $-0.109$ & $0.06$ \\
\midrule
\rowcolor{gray!12}
\multicolumn{4}{@{}c}{\textit{Human reference}} \\
Human majority       & $75$  & $-0.066$ & $0.57$ \\
\bottomrule
\end{tabular}
\end{wraptable}
\textbf{Consequence is a distinct deployment signal.} We measure the Spearman correlation between the task difficulty and the consequence labels produced by each labeling pipeline across SWE-bench Lite
tasks. The three automated pipelines label all 300 tasks, and the human-majority labels cover a 75-task subset. Table~\ref{tab:consequence_orthogonality} shows that consequence has little correlation with task difficulty, as Spearman correlations range from -0.109 to +0.047 across the different pipelines. Appendix~\ref{app:iaa}
shows that the human labels are reliable enough for this comparison, with high annotator self-consistency and moderate agreement with the LLM pipelines, and Appendix~\ref{app:subdomain} shows that the weak relationship is not driven by any single SWE-bench sub-domain. This weak relationship matters for compute allocation, because it means a scheduler cannot infer deployment risk from difficulty alone.  Consequence thus provides a distinct axis of value, but a deployment-time scheduler can use it only if it can be estimated before the model produces a solution, which we test next.
\textbf{Deployment-time consequence is predictable.}
To test whether deployment-time information contains sufficient signal to recover the post-hoc consequence assessment needed for routing, we use a lightweight issue-only predictor to estimate $\hat{s}_x\in\{0,1,2\}$ under the same consequence rubric defined in \S\ref{sec:cons}, using the GitHub issue description and file references. We compare$\hat{s}_x$ against the LLM-with-patch reference label, which has access to post-execution information such as the gold patch and modified files. We use Qwen2.5-7B-Instruct as the primary predictor and Claude Sonnet 4.5 as a cross-model check on SWE-bench Lite, and apply the Qwen predictor to five further pools, namely MSWE-bench mini, a database and ORM subset of SWE-bench Verified, BIRD, FinQA, and MATH (see details in Appendices~\ref{app:crossdataset}, \ref{app:seconddomain}, and~\ref{app:ood}). Table~\ref{tab:predictor} reports Cohen's $\kappa$, the recall and precision of class~2, and how missed class-2 tasks are distributed between classes 1 and 0. 

The primary Qwen predictor reaches $\kappa=0.572$ on SWE-bench Lite, which indicates moderate agreement, and recovers $88.6\%$ of the reference class-2 tasks. More importantly for routing, it predicts none of class-2 tasks as low consequence, so no high-consequence task is sent to the cheapest tier. This property does not depend on a single model family, since the cross-model Claude predictor makes no such error either despite lower overall agreement, and it also transfers across pools. None of the 110 reference class-2 tasks across all six pools is predicted as low consequence, which bounds this error rate below $2.7\%$ at $95\%$ confidence (Appendix~\ref{app:predictor}). The consequence signal is therefore
already present in the task description before solving, and this level of
prediction quality is sufficient for routing: averaged over $1{,}000$ random tiebreaks, predictor-driven routing retains $93.9\%$ of the oracle-consequence gain ($95\%$ CI: $85.9$--$102.5\%$; Appendix~\ref{app:bootstrap}).
\begin{table}[t]
\centering
\caption{\textbf{Deployment-time consequence prediction.}
All predictors use only pre-solution information and are evaluated against
LLM-with-patch reference labels. The key routing failure is underestimating
high-consequence tasks as low consequence ($2\rightarrow0$). Cross-pool
statistics aggregate over Lite, Multi-SWE-bench mini, Verified, BIRD, FinQA,
and MATH (Appendix~\ref{app:predictor}).}
\label{tab:predictor}
\footnotesize
\setlength{\tabcolsep}{2pt}
\begin{adjustbox}{width=0.8\linewidth}
\begin{tabular}{lcccccc}
\toprule
Predictor & Pool & $\kappa$ & Prec$_2$ & Rec$_2$ & $2\rightarrow0$ & $2\rightarrow1$ \\
\midrule

Qwen issue-only
& SWE-bench Lite
& $\mathbf{0.572}$
& $0.46$
& $\mathbf{88.6\%}$
& $\mathbf{0/44}$
& $5/44$ \\

Claude issue-only
& SWE-bench Lite
& $0.379$
& $0.48$
& $52.3\%$
& $\mathbf{0/44}$
& $21/44$ \\

\midrule
\rowcolor{gray!12}
\multicolumn{7}{c}{\textit{Cross-pool transfer (Qwen issue-only predictor)}} \\

Qwen issue-only
& MSWE-bench mini
& $0.382$
& $0.16$
& $100\%$ ($3/3$)
& $\mathbf{0/3}$
& $0/3$ \\

Qwen issue-only
& Verified (DB/ORM)
& $0.582$
& $0.62$
& $95.7\%$ ($45/47$)
& $\mathbf{0/47}$
& $2/47$ \\

Qwen issue-only
& BIRD
& $0.568$
& $1.00$
& $100\%$ ($3/3$)
& $\mathbf{0/3}$
& --- \\

Qwen issue-only
& FinQA
& $0.661$
& $1.00$
& $60\%$ ($6/10$)
& $\mathbf{0/10}$
& --- \\

Qwen issue-only
& MATH
& $0.514$
& $0.00$
& $0\%$ ($0/3$)
& $\mathbf{0/3}$
& --- \\

\midrule
\rowcolor{gray!12}
\multicolumn{7}{c}{\textit{Cross-pool aggregate}} \\

Qwen issue-only
& All 6 pools
& ---
& ---
& ---
& $\mathbf{0/110}$
& --- \\

\bottomrule
\end{tabular}
\end{adjustbox}
\end{table}
\subsection{Main Results: Reallocating Success to High-Consequence Tasks}
\label{sec:main}

\textbf{Experimental Setup} Our primary experiment uses cached per-task outcomes on SWE-bench Lite from 16 public SWE-bench systems spanning multiple model generations and agent
scaffolds. We rank the systems by total resolved count, using the bottom four as
the cheap tier and the top four as the premium tier. 
All strategies operate under the same number of tasks, upgrading $25\%$ of the tasks to the premium tier and leaving the rest on the cheap tier. \textit{Random} selects tasks uniformly at random and is averaged over $1{,}000$ draws. \textit{Difficulty-aware} ranks tasks by the task difficulty or by a deployment-time difficulty forecast from issue text.
\textit{Consequence-aware}, our method, ranks tasks by the predicted weight
$\hat{w}_x$ from the issue-only predictor, with the Qwen predictor as the primary variant, the Claude predictor as a cross-model check, and the with-patch reference label $w_x$ as an oracle. \textit{Priority-aware} ranks
tasks by the exact priority score $\hat{w}_x\Delta_x$ or $w_x\Delta_x$. Because its $\Delta_x$ is computed from observed outcomes, it is reported only as a post-hoc upper bound on what consequence-only routing forgoes. We report the cost-weighted objective $\mathcal{L}$, the success rate on class-2 tasks, and overall accuracy to check that the gain comes from reallocation rather than a uniform improvement. Bootstrap confidence intervals, random-allocation variance, leave-one-class-2-out checks, and split-half estimates for priority-aware variants are reported in Appendix~\ref{app:bootstrap}. Full loss--compute Pareto curves across premium budget fractions are provided in Appendix~\ref{app:pareto}.

\begin{figure}[t]
\centering
\includegraphics[width=0.95\linewidth]{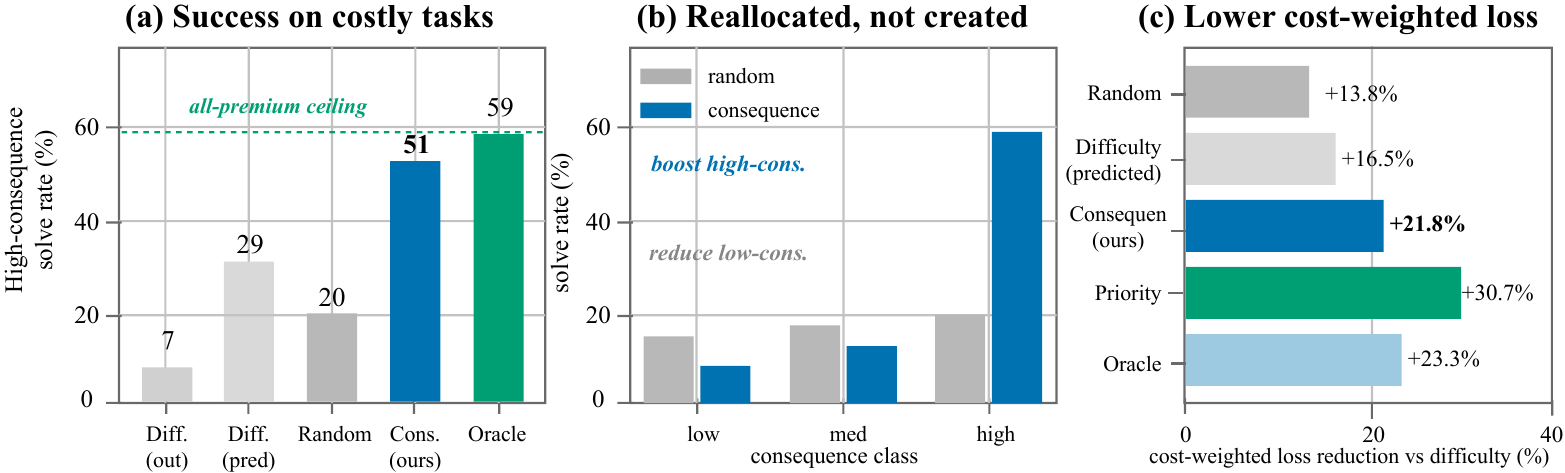}
\caption{\textbf{Consequence-aware routing reallocates success toward high-consequence tasks.}
(a) Under the same budget, our strategy yields $51.1\%$ class-2 success.
(b) Gains arise from redistribution across consequence classes rather
than a uniform accuracy improvement.
(c) Consequence-aware routing reduces cost-weighted loss, while priority-aware
routing is a post-hoc upper bound.}
\label{fig:main}
\end{figure}
\begin{table}[t]
\centering
\caption{\textbf{Main routing results on SWE-bench Lite.}
Consequence-aware routing reduces cost-weighted loss by
reallocating success toward high-consequence tasks. Priority-aware strategies are not applicable for deployment due to the use of empirical marginal gain. Reduced loss is relative to outcome-derived difficulty.}
\label{tab:main}
\large
\renewcommand{\arraystretch}{1.05}
\setlength{\tabcolsep}{10pt}
\begin{adjustbox}{width=0.95\linewidth}
\begin{tabular}{@{}lrrrr@{}}
\toprule
Strategy
& $\mathcal{L}$
& Reduced Loss 
& Class-2 succ.
& Acc. \\
\midrule

\rowcolor{gray!12}
\multicolumn{5}{c}{\textit{Difficulty-based and random baselines}} \\[2pt]

Difficulty-aware (outcome-derived)
& $268.25$
& $0\%$
& $6.8\%$
& $0.054$ \\

Difficulty-aware (deployment predicted)
& $224.00$
& $+16.5\%$
& $29.0\%$
& $0.181$ \\

Random (mean over $1{,}000$ draws)
& $231.20$
& $+13.8\%$
& $19.4\%$
& $0.180$ \\

\midrule
\rowcolor{gray!12}
\multicolumn{5}{c}{\textit{Consequence-aware routing}} \\[2pt]

Consequence-aware (Claude issue-only)
& $220.50$
& $+17.8\%$
& $36.9\%$
& $0.174$ \\

\textbf{Consequence-aware (Qwen issue-only; deployment-time)}
& $209.75$
& $+21.8\%$
& $\mathbf{51.1\%}$
& $0.184$ \\

Consequence-aware (with-patch oracle)
& $205.75$
& $+23.3\%$
& $58.5\%$
& $0.187$ \\

\midrule
\rowcolor{gray!12}
\multicolumn{5}{c}{\textit{Priority-aware analysis}} \\[2pt]

Priority-aware (Qwen consequence $\times$ marginal gain)
& $186.00$
& $+30.7\%$
& $50.6\%$
& $0.274$ \\

Priority-aware (oracle consequence $\times$ marginal gain)
& $\mathbf{179.75}$
& $\mathbf{+33.0\%}$
& $55.1\%$
& $\mathbf{0.278}$ \\

\bottomrule
\end{tabular}
\end{adjustbox}
\end{table}

\textbf{Consequence-aware routing improves success on high-consequence tasks.}
As shown in Fig.~\ref{fig:main} and Table~\ref{tab:main}, the
deployment-time Qwen consequence router reduces cost-weighted loss by $21.8\%$ relative to outcome-derived difficulty routing. The gain is concentrated on class-2 tasks: our Qwen router achieves $51.1\%$ class-2
success, compared with $29.0\%$ for predicted difficulty and $19.4\%$ for random routing, while overall accuracy remains nearly unchanged. Consequence-aware routing thus improves success where errors are costly rather than uniformly increasing accuracy, and
Appendix~\ref{app:bootstrap} further shows that this improvement is robust to bootstrap resampling and remains stable when each class-2 task is removed in turn.

\textbf{The deployment-time predictor is effective for routing.}
Our issue-only router achieves a cost-weighted loss of $209.75$, close to the oracle's $205.75$. Across $1{,}000$ random tiebreaks, it retains $93.9\%$ of the oracle-consequence gain over outcome-derived difficulty (Appendix~\ref{app:bootstrap}). Note that the with-patch oracle requires information unavailable at deployment time, while the issue-only predictor only uses the issue text and pre-solution file references. The downstream routing result confirms that the predictor quality in \S\ref{sec:signal} can support effective allocation, since routing needs a ranking that keeps high-consequence tasks out of the cheap tier rather than exact labels. 

\textbf{Empirical marginal gain provides an upper-bound-style improvement.}
As expected from Eq.~\ref{eq:budgetobj}, priority-aware routing further reduces
cost-weighted loss by combining consequence with empirical marginal gain
$\Delta_x$. Using predicted and oracle consequence, it achieves $30.7\%$ and
$33.0\%$ loss reductions, respectively. Notably, these variants do not improve
class-2 success over consequence-only routing, showing that lower weighted loss
does not necessarily imply higher success on the highest-consequence class.

These results should be viewed as post-hoc upper-bound-style analyses rather
than deployable methods, since they require empirical $\Delta_x$ computed from
observed outcomes. Appendix~\ref{app:marginal} shows that $\Delta_x$ cannot be
predicted reliably before execution, while Appendix~\ref{app:bootstrap} shows
that using the same outcomes for selection and evaluation makes the empirical
priority results optimistic by roughly $6$--$7$ percentage points. Reliable
marginal-gain estimates could therefore further improve consequence-aware
routing, but obtaining such estimates at deployment time remains an open
challenge.

\subsection{Controlled Within-Model Allocation}
\label{sec:singlemodel}
In the main experiment, upgrading a task means routing it to a stronger system,
so the premium tier differs from the cheap tier in model capability as well as
compute. To test whether the gains come from compute allocation itself, we
repeat the comparison with a single Claude Sonnet~4.5 agent under a fixed prompt
and thinking budget, varying only the number of independent attempts per task. A task is
considered solved if any attempt passes its tests. The number of attempts provides
a meaningful compute lever, with pass rate increasing from
$\mathrm{pass@}1=0.142$ to $\mathrm{pass@}8=0.370$, whereas varying the
thinking-token budget yields no consistent improvement
(Appendix~\ref{app:within}). Each strategy assigns eight attempts to selected tasks and one to the rest under the same total attempt budget, so they differ only in which tasks receive extra compute.

Table~\ref{tab:withinmain} shows that the main result persists when the model is fixed. Consequence-aware routing reduces cost-weighted loss by $18.6\%$ relative to difficulty-aware routing, with a bootstrap interval of $[+10.5,+30.7]$, and also outperforms random allocation. Priority-aware routing further improves performance, consistent with the benefit of marginal-gain information. The issue-only predictor slightly outperforms the with-patch oracle because of discrete allocation and ties, rather than better label quality. A second evaluation with a stricter DeepSeek judge shows the same trend: consequence-aware and priority-aware routing still outperform difficulty-aware routing, although the gains are smaller (Appendix~\ref{app:within}).


\begin{table}[t]
\centering
\caption{\textbf{Controlled within-model allocation.}
Using a single fixed Claude Sonnet~4.5 agent, compute is varied only
through the number of best-of-$k$ attempts. The same strategy ordering
holds when model identity and prompt are fixed.}
\label{tab:withinmain}
\large
\renewcommand{\arraystretch}{0.9}
\setlength{\tabcolsep}{10pt}
\begin{adjustbox}{width=0.95\linewidth}
\begin{tabular}{@{}lrrc@{}}
\toprule
Strategy
& $\mathcal{L}$
& $\Delta$ vs.\ Diff
& Bootstrap CI \\
\midrule
\rowcolor{gray!12}
\multicolumn{4}{c}{\textit{Baselines}} \\

Difficulty-aware
& $102.00$
& $0\%$ baseline
& --- \\

Random (mean over $1{,}000$ draws)
& $91.92$
& $+9.9\%$
& --- \\

\midrule
\rowcolor{gray!12}
\multicolumn{4}{c}{\textit{Consequence-aware allocation}} \\

\textbf{Consequence-aware (issue-only predictor)}
& $83.00$
& $+18.6\%$
& $[+10.5, +30.7]$ \\

Consequence-aware (with-patch oracle)
& $85.00$
& $+16.7\%$
& --- \\

\midrule
\rowcolor{gray!12}
\multicolumn{4}{c}{\textit{Priority-aware analysis}} \\

Priority-aware (predictor $\times$ gain)
& $\mathbf{77.00}$
& $\mathbf{+24.5\%}$
& $\mathbf{[+14.6, +35.6]}$ \\

\bottomrule
\end{tabular}
\end{adjustbox}

\end{table}

\subsection{Robustness, replication, and boundary conditions.}
\label{sec:boundary}

Eq.~\eqref{eq:budgetobj} shows that consequence-only routing is most useful when costs vary across tasks, premium compute provides nonzero gains, and difficulty does not already identify high-$\Delta_x$ tasks. Since SWE-bench Lite satisfies these conditions, we test whether the observed gains persist when the data, cost model, and judge change, and when these conditions no longer hold.

\textbf{The result is robust across software settings.} We repeat the comparison on 140 database and ORM tasks from SWE-bench Verified
that do not overlap with SWE-bench Lite, building the cheap and premium tiers
from a separate set of 20 public systems. Consequence-aware routing again
improves on both random and difficulty-aware routing, and priority-aware routing
remains best, although the margin over random is smaller because prediction
error costs more on this pool (Appendix~\ref{app:seconddomain}). The main observations also hold when the cost model changes, since it is preserved under five
monotone mappings from consequence classes to scalar weights and under
continuous severity scores elicited from two judge families
(Appendix~\ref{app:costmap}), as well as when each labeling pipeline supplies
both the routing signal and the evaluation weights
(Appendix~\ref{app:rubric}).

\begin{figure}[ht]
\centering
\includegraphics[width=0.95\linewidth]{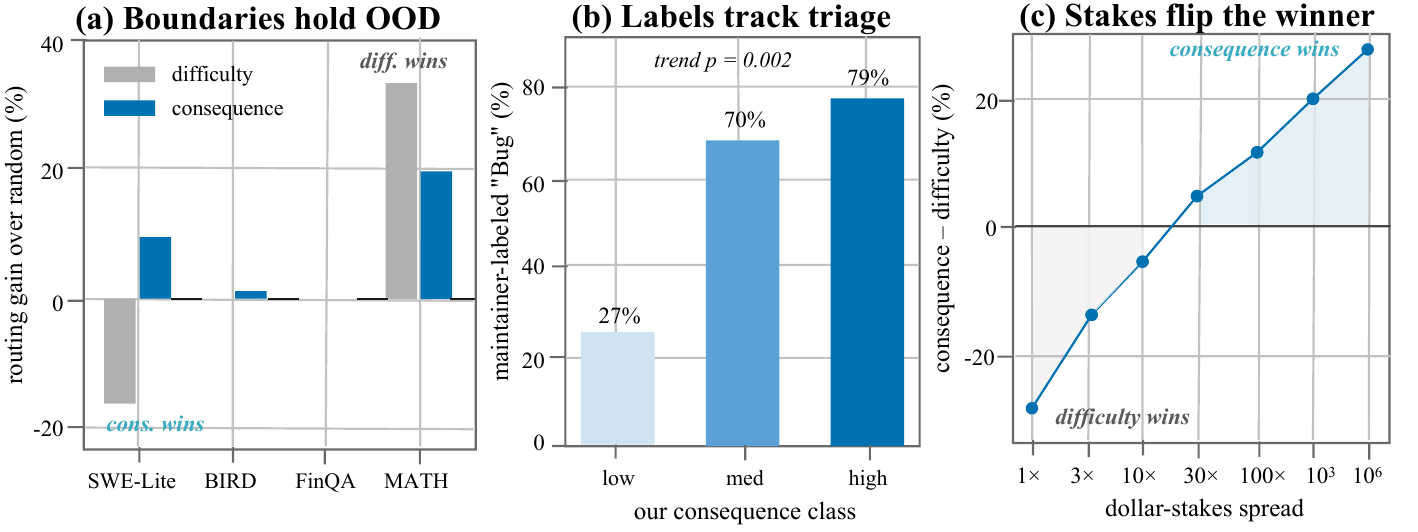}
\caption{\textbf{Boundary conditions and external validation.}
(a) Routing performance varies across domains.
(b) Consequence labels show consistent trends with maintainer triage behavior.
(c) Increasing the error-cost spread on MATH shifts the preferred strategy
from difficulty-aware to consequence-aware routing.}
\label{fig:generalization}
\end{figure}

\textbf{The preferred routing signal depends on cost and compute structure.}
Figure~\ref{fig:generalization} summarizes the main boundary conditions. On BIRD, consequence costs are nearly homogeneous; on FinQA, premium compute
provides little additional benefit; and on MATH, difficulty is positively correlated with marginal gain ($\rho=+0.257$, $p=0.002$), making
difficulty-aware routing more effective. Increasing the high-to-low error-cost spread on MATH reverses this comparison at roughly $30\times$
(Figure~\ref{fig:generalization}(c)), while changing the declared deployment context also changes which SWE-bench Lite tasks receive premium compute (Appendix~\ref{app:context}). Overall, consequence-aware routing is most useful when error costs are heterogeneous, premium compute has useful headroom, and difficulty is not already aligned with marginal gain.

\section{Discussion and Limitations}
\label{sec:discussion}

A central limitation is that consequence is estimated rather than observed as realized deployment harm. Our deployment-time scheduler reduces this concern because the issue-only predictor never accesses gold patches, generated fixes, or execution outcomes, while human annotations, alternative labeling pipelines, maintainer behavior, and
cost-mapping analyses provide complementary checks on the construct (Appendices~\ref{app:iaa}--\ref{app:judge2}). These validations support
consequence as a deployment-relevant signal, but do not measure the true downstream cost of an error in production.

Following this, performance depends on how well consequence can be estimated from pre-solution inputs. More fundamentally, the exact priority
rule depends on both $w_x$ and $\Delta_x$, while the estimators we tested do
not predict $\Delta_x$ reliably (Appendix~\ref{app:marginal}). This motivates
the current consequence-only scheduler, while the priority-aware results suggest
that reliable gain estimates, potentially obtained from online execution or
verification feedback, could further improve allocation.

Finally, consequence-aware routing is not always preferable to
difficulty-aware allocation, as its advantage depends on the cost structure and
the additional compute gain, detailed in
\S\ref{sec:boundary}. Our strongest evidence is from software-engineering
tasks, where consequence heterogeneity and matched-budget evaluation are readily
observable, and broader deployment settings remain open. Future work should incorporate richer deployment context, learned cost models, and adaptive policies that combine consequence prediction with online estimates of marginal
compute benefit.

\section{Conclusion}
\label{sec:conclusion}

Adaptive test-time compute is usually routed by difficulty, uncertainty, or expected accuracy gain. This paper argues that deployment requires an additional
routing dimension, namely the consequence of being wrong. We formulated allocation as a cost-weighted scheduling problem and showed that the optimal priority is consequence multiplied by marginal compute gain. Since the latter cannot be estimated before execution, our scheduler ranks tasks by predicted consequence alone, leaving the base reasoning model unchanged. On SWE-bench Lite, consequence-aware routing reallocates success toward costly
failures under a matched budget, substantially improving high-consequence task
success while overall accuracy remains nearly unchanged. A controlled within-model experiment reproduces the same effect when only best-of-$k$ attempts are reallocated. The broader lesson is that test-time compute should not be allocated only by where a model is likely to fail, because
in deployment the value of extra reasoning also depends on what happens when it
fails. Consequence is therefore not a replacement for difficulty or
marginal-gain estimation, but the missing cost term that determines where
successful reasoning matters most.
\bibliography{references}
\bibliographystyle{plainnat}

\appendix

\clearpage

\noindent{\large\scshape Appendix Contents}

\vspace{1.0em}
\small

\newcommand{\appcontentsline}[3]{%
  \noindent
  \makebox[2em][l]{#1}%
  #2\ \dotfill\ \pageref{#3}\par
  \vspace{1em}
}

\appcontentsline{A}
{Full Audit of Native Thinking Allocation}
{app:audit}

\appcontentsline{B}
{Human-Label Robustness Check}
{app:iaa}

\appcontentsline{C}
{Rubric Ablation: All Labeling Pipelines Preserve the Conclusion}
{app:rubric}

\appcontentsline{D}
{External Validity: Maintainer-Revealed Severity}
{app:realcost}

\appcontentsline{E}
{Statistical Robustness of Allocation Results}
{app:bootstrap}

\appcontentsline{F}
{Pareto Curves Across Compute Budgets}
{app:pareto}

\appcontentsline{G}
{Cost-Mapping and Continuous-Severity Sensitivity}
{app:costmap}

\appcontentsline{H}
{Predictor Calibration Details}
{app:predictor}

\appcontentsline{I}
{Per-Task Marginal Utility and Marginal-Gain Prediction}
{app:marginal}

\appcontentsline{J}
{Why Difficulty-Aware Allocation Fails}
{app:whyfail}

\appcontentsline{K}
{Robustness Across SWE-bench Sub-Domains}
{app:subdomain}

\appcontentsline{L}
{Cross-Dataset Transfer: Multi-SWE-bench mini}
{app:crossdataset}

\appcontentsline{M}
{A Second Software Pool: Database/ORM Tasks from SWE-bench Verified}
{app:seconddomain}

\appcontentsline{N}
{Judge Robustness for Consequence Labels}
{app:judge2}

\appcontentsline{O}
{Out-of-Domain Probes: BIRD, FinQA, and MATH}
{app:ood}

\appcontentsline{P}
{Within-Model Allocation: Attempts as the Compute Lever}
{app:within}

\appcontentsline{Q}
{A Live Online Deployment Run}
{app:online}

\appcontentsline{R}
{Context-Conditional Consequence}
{app:context}

\appcontentsline{S}
{Deployment-Time Consequence-Only Scheduler}
{app:scheduler}

\normalsize
\clearpage

\section{Full Audit of Native Thinking Allocation}
\label{app:audit}

This appendix provides the full per-model audit behind
Figure~\ref{fig:motivation}(a) and Table~\ref{tab:audit-app}. The main text
reports the central finding: native reasoning computation is not reliably
aligned with task consequence. Here we provide model-specific measurement
details and saturation statistics.

\paragraph{Qwen3-8B.}
Qwen3-8B is evaluated on all $300$ SWE-bench Lite tasks with
\texttt{max\_new\_tokens}$=4096$. Thinking length varies across tasks,
but its rank correlation with consequence is essentially zero:
$\rho=+0.002$ ($p=0.97$). This is the cleanest ``no signal'' case.

\paragraph{Qwen3-VL-8B-Thinking.}
Qwen3-VL-8B-Thinking reaches the configured token cap on almost every
task: $298/300$ tasks ($99.3\%$) at an $8192$-token cap and
$299/300$ tasks ($99.7\%$) at a $4096$-token cap. With almost all
tasks pinned at the cap, the model has no effective headroom for
task-specific consequence allocation.

\paragraph{Claude Sonnet 4.5.}
Claude Sonnet 4.5 with extended thinking is evaluated on a frozen
$171$-task stratified SWE-bench Lite sample with a $16000$-token
thinking budget. Thinking length has a statistically significant but
small positive correlation with consequence:
$\rho=+0.203$ ($p=0.008$). High-consequence tasks receive $20.5\%$
more thinking than low-consequence tasks on average.

\paragraph{o4-mini.}
OpenAI's o4-mini is evaluated on the same $171$-task sample. Since the
API reports reasoning-token counts, we use those counts as the
per-task compute measure. The correlation is again positive but small:
$\rho=+0.239$ ($p=0.002$), with class-$2$ tasks receiving $31.4\%$
more reasoning tokens than class-$0$ tasks.

\paragraph{DeepSeek-R1.}
DeepSeek-R1 returns an observable reasoning stream. On the same
$171$-task sample, it exhibits a saturation failure mode at frontier
scale: $92\%$ of tasks truncate under an $8192$-token cap, and $54\%$
still truncate under a $24000$-token cap. Among tasks that terminate
naturally, the remaining consequence trend is weak and not
statistically significant.

\paragraph{Interpretation.}
The audit does not show that thinking computation is useless. It shows
that native thinking length is not reliably aligned with the
consequence-weighted objective. This motivates the scheduler-level
intervention of \S\ref{sec:method}: rather than assuming
consequence-awareness emerges from the model's native thinking process,
we add it explicitly at the allocation layer.
\begin{table}[H]
\centering
\caption{\textbf{Native thinking allocation audit on SWE-bench Lite.}
We report sample size, configured token budget, Spearman correlation
between thinking length and consequence, and the observed failure mode.
The audited models exhibit no consequence signal, saturation, or only
weak adaptation.}
\label{tab:audit-app}

\small
\renewcommand{\arraystretch}{1.10}
\setlength{\tabcolsep}{5pt}

\begin{adjustbox}{width=\linewidth}
\begin{tabular}{@{}lrr@{\hspace{8pt}}ccc@{\hspace{8pt}}l@{}}
\toprule
&
\multicolumn{2}{c}{\textbf{Audit setup}}
&
\multicolumn{3}{c}{\textbf{Consequence sensitivity}}
&
\\
\cmidrule(lr){2-3}
\cmidrule(lr){4-6}

Model
& $n$
& Max new tokens
& $\rho(\mathrm{thinking},\mathrm{cons.})$
& $p$
& High vs.\ Low
& Failure mode \\
\midrule

\rowcolor{gray!12}
\multicolumn{7}{c}{\textit{Absent or weak consequence adaptation}} \\

Qwen3-8B (hybrid)
& $300$
& $4{,}096$
& $+0.002$
& n.s.
& ---
& no signal \\

Claude Sonnet 4.5 (ext.\ think)
& $171$
& $16{,}000$
& $+0.203$
& $0.008$
& $+20.5\%$
& weak but inadequate \\

o4-mini (reasoning API)
& $171$
& $12{,}000$
& $+0.239$
& $0.002$
& $+31.4\%$
& weak but inadequate \\

\midrule
\rowcolor{gray!12}
\multicolumn{7}{c}{\textit{Thinking-length saturation}} \\

Qwen3-VL-8B-Thinking
& $300$
& $8{,}192$
& ---
& ---
& ---
& $\mathbf{99.3\%}$ saturated \\

DeepSeek-R1 (open CoT API)
& $171$
& $24{,}000$
& ---
& ---
& ---
& $\mathbf{54\%}$ saturated \\

\bottomrule
\end{tabular}
\end{adjustbox}

\end{table}
\section{Human-Label Robustness Check}
\label{app:iaa}

This appendix validates the consequence construct with human
annotations. The goal is not to replace the full-pool LLM-with-patch
labels, but to test whether the central patterns of the paper survive
under human-majority labels.

\paragraph{Study design.}
We sampled $150$ unique tasks, stratified by the LLM-with-patch
consequence label: $75$ from SWE-bench Lite and $75$ from
Multi-SWE-bench mini. Each annotator received $165$ rows in randomized
order, including $15$ hidden duplicates, five per class, to measure
self-consistency. Three annotators participated: one author, one CS PhD
outside the project, and one non-author CS graduate student. Annotators
did not see the gold patch, file diff, or one another's labels. Each
annotator first completed a $9$-task calibration warm-up with reference
labels.

\paragraph{Self-consistency.}
Cohen's $\kappa$ on the hidden duplicates is $1.000$ for two
annotators and $0.789$ for the third. Raw duplicate agreement is
$100\%$, $100\%$, and $86.7\%$, respectively. The rubric is therefore
internally stable at the annotator level.

\paragraph{Human labels preserve consequence--difficulty orthogonality.}
On the $75$ SWE-bench Lite tasks in the human study, the Spearman
correlation between human-majority consequence and difficulty is
\[
\rho(\mathrm{human\mbox{-}cons},\mathrm{difficulty})=-0.066,
\qquad p=0.57.
\]
This reproduces the main-text conclusion that consequence is not
difficulty. At $n=75$, the study is powered only against correlations
larger than roughly $|\rho|\gtrsim 0.23$, so we interpret this as a
construct-validity check rather than a precise full-pool estimate.

\paragraph{Agreement with LLM labels.}
Table~\ref{tab:human-vs-llm} compares the human-majority labels with
the three labeling pipelines used in the paper.
\begin{table}[h]
\centering
\caption{\textbf{Agreement with human-majority consequence labels.}
Cohen's $\kappa$ between human-majority labels and each labeling pipeline
on the $75$ SWE-bench Lite tasks included in the human study.}
\label{tab:human-vs-llm}

\small
\renewcommand{\arraystretch}{1.12}

\begin{tabular*}{\linewidth}{@{\extracolsep{\fill}}lc@{}}
\toprule
Pipeline
& $\kappa$ vs.\ human majority \\
\midrule

\rowcolor{gray!12}
\multicolumn{2}{c}{\textit{Rule-based labeling}} \\

Rule-based labeler
& $0.223$ \\

\addlinespace[2pt]
\rowcolor{gray!12}
\multicolumn{2}{c}{\textit{LLM-based labeling}} \\

LLM-with-patch judge (Qwen2.5-7B-Instruct)
& $0.500$ \\

LLM issue-only predictor (Qwen2.5-7B-Instruct)
& $\mathbf{0.517}$ \\

\bottomrule
\end{tabular*}

\end{table}
The two LLM pipelines show moderate agreement with the human majority,
while the rule-based pipeline is substantially weaker. This supports
using the LLM-with-patch judge as the full-pool primary label, while
treating the human labels as a construct-validity check.

\paragraph{Final human label distribution.}
Of the $150$ tasks, $142$ have a clear majority vote. The remaining
$8$ are three-way splits and are conservatively assigned to the middle
class. The final human-majority distribution is
$\{0{:}45,\ 1{:}67,\ 2{:}38\}$.

\section{Rubric Ablation: All Labeling Pipelines Preserve the Conclusion}
\label{app:rubric}

This appendix checks whether the central findings depend on the
specific LLM-with-patch labeling pipeline used in the main text. We
rerun two analyses under four labelings: rule-based, LLM-with-patch,
LLM issue-only, and human-majority labels.

\paragraph{Orthogonality under all labelings.}
Table~\ref{tab:abl-ortho} reports the Spearman correlation between
consequence and difficulty. All correlations are within $\pm0.11$ of
zero and none is significant at $\alpha=0.05$.

\begin{table}[h]
\centering
\caption{\textbf{Rubric ablation: consequence--difficulty orthogonality.}
Spearman correlations between consequence and task difficulty remain small
across automated and human labeling pipelines.}
\label{tab:abl-ortho}

\small
\renewcommand{\arraystretch}{1.12}

\begin{tabular*}{\linewidth}{@{\extracolsep{\fill}}lccc@{}}
\toprule
Labeling pipeline
& $n$
& Spearman $\rho$
& $p$-value \\
\midrule

\rowcolor{gray!12}
\multicolumn{4}{c}{\textit{Automated labeling}} \\

Rule-based
& $300$
& $+0.047$
& $0.42$ \\

LLM-with-patch (paper primary)
& $300$
& $-0.096$
& $0.10$ \\

LLM issue-only predictor
& $300$
& $-0.109$
& $0.06$ \\

\addlinespace[2pt]
\rowcolor{gray!12}
\multicolumn{4}{c}{\textit{Human reference}} \\

Human majority
& $75$
& $-0.066$
& $0.57$ \\

\bottomrule
\end{tabular*}

\end{table}
\paragraph{Allocation gain under all labelings.}
For each labeling, we recompute the matched-compute allocation
comparison, using that labeling both as the oracle routing signal and
as the cost weight in the loss. Table~\ref{tab:abl-gain} shows that
priority-aware oracle routing improves over difficulty-aware routing
by $30\%+$ under every labeling.
\begin{table}[h]
\centering
\caption{\textbf{Rubric ablation: allocation gain.}
Cost-weighted loss at the matched top-$25\%$ premium-compute budget,
recomputed under each consequence labeling pipeline. For each row, the same
labeling defines both the consequence signal and the loss weights.}
\label{tab:abl-gain}

\small
\renewcommand{\arraystretch}{1.10}
\setlength{\tabcolsep}{4.5pt}

\begin{adjustbox}{width=\linewidth}
\begin{tabular}{@{}lr@{\hspace{6pt}}rrrr@{\hspace{7pt}}r@{}}
\toprule
&
&
\multicolumn{4}{c}{\textbf{Cost-weighted loss $\mathcal{L}$}}
&
\multicolumn{1}{c}{\textbf{Gain}} \\
\cmidrule(lr){3-6}
\cmidrule(l){7-7}

Consequence labeling
& $n$
& Diff.
& Random
& Cons.-oracle
& Priority-oracle
& $\Delta$ vs.\ Diff. \\
\midrule

\rowcolor{gray!12}
\multicolumn{7}{c}{\textit{Automated labeling}} \\

Rule-based
& $300$
& $406.2$
& $352.5$
& $330.8$
& $\mathbf{281.2}$
& $+30.8\%$ \\

\textbf{LLM-with-patch (paper primary)}
& $300$
& $268.2$
& $233.8$
& $205.8$
& $\mathbf{179.8}$
& $+33.0\%$ \\

LLM issue-only
& $300$
& $307.5$
& $266.2$
& $229.5$
& $\mathbf{201.5}$
& $+34.5\%$ \\

\addlinespace[2pt]
\rowcolor{gray!12}
\multicolumn{7}{c}{\textit{Human reference}} \\

Human majority
& $75$
& $75.8$
& $67.5$
& $55.2$
& $\mathbf{48.2}$
& $+36.3\%$ \\

\bottomrule
\end{tabular}
\end{adjustbox}

\end{table}
The conclusion is stable: the consequence construct is not an artifact
of one LLM judge or one rubric implementation.

\section{External Validity: Maintainer-Revealed Severity}
\label{app:realcost}

Figure~\ref{fig:generalization}(b) summarizes the main
external-validity result. This appendix provides the mining procedure,
repository-level checks, null process signals, and additional
maintainer metadata analyses.

\paragraph{Backport mining.}
Each SWE-bench Lite instance encodes its originating pull request. For
$297/300$ tasks with resolvable merge commits, we mined the
corresponding repositories for whether the fix was backported. We also
recorded PR open-to-merge duration, discussion intensity, and security
markers.

\paragraph{Backporting tracks consequence.}
Backport rate increases monotonically with consequence:
\[
8.3\% \ (5/60) \quad \text{for class }0,
\]
\[
20.6\% \ (40/194) \quad \text{for class }1,
\]
\[
41.9\% \ (18/43) \quad \text{for class }2.
\]
The ordered trend is significant ($\rho=+0.232$, $p=0.0001$), and the
class-$2$ vs rest odds ratio is $3.34$ (Fisher $p=0.0009$). Part of
the pooled association may reflect repository composition, so we also
report stratified checks. A Cochran--Mantel--Haenszel test within
repositories gives pooled OR $1.76$ ($p=0.14$), and restricting to
repositories that backport at all preserves the monotone gradient
($19.2\%/36.3\%/47.4\%$) with an ordered trend
$\rho=+0.175$ ($p=0.024$). Thus, the pooled signal is not solely a
repository-mix artifact, although the repository-stratified contrast is
underpowered.

\paragraph{Null signals.}
PR open-to-merge duration does not separate consequence classes
(medians $2.0/1.9/1.6$ days; Kruskal--Wallis $p=0.56$). Comment
intensity also does not separate classes (medians $4/5/3$;
$p=0.19$). Security markers are too rare in SWE-bench Lite to be
informative. These nulls are useful: they suggest that backporting
captures severity more directly than general process friction.

\paragraph{Django Trac metadata.}
For the $114$ Django tasks, issue-tracker metadata provides an
LLM-independent anchor. Maintainer-assigned issue type tracks the
consequence label: the fraction labeled as genuine ``Bug'' rises from
$26.7\%$ at class~0 to $69.7\%$ at class~1 and $78.8\%$ at class~2
(ordered trend $\rho=+0.282$, $p=0.002$). The ``easy pickings''
difficulty flag is approximately orthogonal to consequence
($\rho=-0.12$, n.s.). Thus, maintainer behavior supports the paper's
central separation between consequence and difficulty.

\section{Statistical Robustness of Allocation Results}
\label{app:bootstrap}

This appendix quantifies the stability of Table~\ref{tab:main}. We
report tiebreak variance, split-half priority estimates, task-level
bootstrap intervals, random-allocation variance, and a
leave-one-class-2-out analysis.

\paragraph{Tiebreak variance.}
Ordinal consequence scores create ties. Re-drawing the tiebreak vector
$1{,}000$ times gives predictor loss $210.2\pm1.8$ with $95\%$
interval $[206.8,213.5]$ and oracle-consequence loss $206.4\pm2.0$.
The headline retention of oracle-consequence gain has mean $93.9\%$
and $95\%$ interval $[85.9\%,102.5\%]$.

\paragraph{Split-half priority estimates.}
Priority-aware rows select tasks using empirical marginal gain
estimated from the same tier outcomes used for evaluation. This is
optimistic because of selection on noise. Splitting each tier in half,
selecting on one half and evaluating on the other over $200$ random
splits, gives $+27.2\%\pm1.6$ for oracle priority, compared with
$+33.8\%$ when selecting and evaluating on the same half. The priority
rows should therefore be read as optimistic by roughly $6$--$7$ points.

\paragraph{Task-level bootstrap.}
We resample the $300$ SWE-bench Lite tasks with replacement $5{,}000$
times. Table~\ref{tab:bootstrap} reports percentile intervals.

\begin{table}[h]
\centering
\caption{\textbf{Task-level bootstrap.}
Results are based on $5{,}000$ task-level resamples at the top-$25\%$
matched-compute operating point.}
\label{tab:bootstrap}

\small
\renewcommand{\arraystretch}{1.10}
\setlength{\tabcolsep}{5pt}

\begin{adjustbox}{width=\linewidth}
\begin{tabular}{@{}lrr@{\hspace{8pt}}rr@{}}
\toprule
&
\multicolumn{2}{c}{\textbf{Cost-weighted loss}}
&
\multicolumn{2}{c}{\textbf{Gain vs.\ difficulty}} \\
\cmidrule(lr){2-3}
\cmidrule(l){4-5}

Strategy
& $\mathcal{L}$
& $95\%$ CI
& $\Delta$ vs.\ Diff.
& $95\%$ CI \\
\midrule

\rowcolor{gray!12}
\multicolumn{5}{c}{\textit{Baselines}} \\

Difficulty-aware
& $268.25$
& $[249.2,286.8]$
& baseline
& --- \\

Random
& $231.20$
& $[214.8,251.2]$
& $+13.8\%$
& $[+9.9,+16.2]\%$ \\

\midrule
\rowcolor{gray!12}
\multicolumn{5}{c}{\textit{Consequence-aware routing}} \\

Consequence-aware (Claude issue-only)
& $220.50$
& $[201.2,239.8]$
& $+17.8\%$
& $[+14.1,+21.5]\%$ \\

\textbf{Consequence-aware (Qwen issue-only)}
& $209.75$
& $[190.8,230.0]$
& $\mathbf{+21.8\%}$
& $\mathbf{[+17.6,+25.5]\%}$ \\

Consequence-aware (with-patch oracle)
& $205.75$
& $[187.0,223.8]$
& $+23.3\%$
& $[+19.2,+27.4]\%$ \\

\midrule
\rowcolor{gray!12}
\multicolumn{5}{c}{\textit{Priority-aware analysis}} \\

Priority-aware (Qwen + gain)
& $186.00$
& $[169.5,203.0]$
& $+30.7\%$
& $[+27.7,+33.4]\%$ \\

Priority-aware (oracle + gain)
& $\mathbf{179.75}$
& $\mathbf{[163.0,197.5]}$
& $\mathbf{+33.0\%}$
& $\mathbf{[+30.3,+35.4]\%}$ \\

\bottomrule
\end{tabular}
\end{adjustbox}

\end{table}
\paragraph{Random allocation variance.}
At the fixed task set, $1{,}000$ independent random top-$25\%$
allocations have mean loss $231.20$, standard deviation $3.70$, and
$95\%$ interval $[224.0,238.2]$. All $1{,}000$ random allocations
outperform outcome-derived difficulty routing.

\paragraph{Leave-one-class-2-out.}
Removing each of the $44$ high-consequence tasks in turn, the Qwen
issue-only consequence router remains within $[+21.5,+22.3]\%$
reduction over difficulty-aware routing and wins in all $44$ runs. The
priority-aware row remains within $[+30.3,+30.9]\%$ and also wins in
all $44$ runs.

\section{Pareto Curves Across Compute Budgets}
\label{app:pareto}
Figure~\ref{fig:main}(c) reports the matched top-$25\%$ operating
point. This appendix provides the full loss--compute Pareto sweep over
premium budget fractions.
\begin{figure}[h]
\centering
\includegraphics[width=\linewidth]{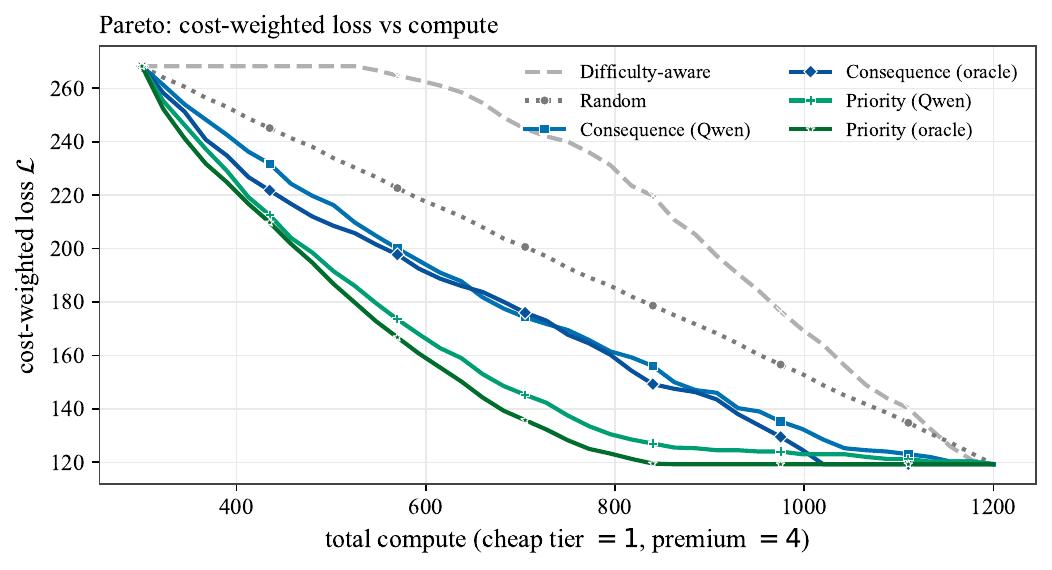}
\caption{\textbf{Cost-weighted loss vs total compute on the
16-system SWE-bench compute-tier benchmark.} Each curve sweeps the
fraction of tasks routed to the premium tier from $0$ to $100\%$.
Consequence-aware routing improves over difficulty-aware routing
across the operating region, while priority-aware variants provide the
best analysis upper bound when empirical marginal gain is available.
The dotted vertical line marks the top-$25\%$ premium operating point
used in Table~\ref{tab:main}.}
\label{fig:pareto-app}
\end{figure}

\section{Cost-Mapping and Continuous-Severity Sensitivity}
\label{app:costmap}

The main text maps consequence classes $\{0,1,2\}$ directly to scalar
weights. This appendix tests whether the results depend on that
particular mapping.

\paragraph{Monotone mappings.}
Table~\ref{tab:costmap} recomputes the matched-compute comparison
under five monotone cost mappings. The ordering
Consequence-aware $>$ Random $>$ Difficulty-aware and
Priority-aware $>$ Consequence-aware is preserved under every mapping.

\begin{table}[h]
\centering
\caption{\textbf{Cost-mapping sensitivity.}
Cost-weighted loss under alternative mappings from consequence classes to
scalar weights. Parentheses report the relative reduction in loss versus
difficulty-aware routing.}
\label{tab:costmap}

\small
\renewcommand{\arraystretch}{1.10}
\setlength{\tabcolsep}{4pt}

\begin{adjustbox}{width=\linewidth}
\begin{tabular}{@{}lrr@{\hspace{6pt}}rr@{\hspace{6pt}}rr@{}}
\toprule
&
\multicolumn{2}{c}{\textbf{Baselines}}
&
\multicolumn{2}{c}{\textbf{Consequence-aware}}
&
\multicolumn{2}{c}{\textbf{Priority-aware}} \\
\cmidrule(lr){2-3}
\cmidrule(lr){4-5}
\cmidrule(l){6-7}

Mapping
& Diff.
& Random
& Qwen
& Oracle
& Qwen
& Oracle \\
\midrule

\rowcolor{gray!12}
\multicolumn{7}{c}{\textit{Main mapping}} \\

$\{0,1,2\}$
& $268.2$
& $231.2$
& $\mathbf{209.8\ (+21.8\%)}$
& $205.8\ (+23.3\%)$
& $186.0\ (+30.7\%)$
& $179.8\ (+33.0\%)$ \\

\midrule
\rowcolor{gray!12}
\multicolumn{7}{c}{\textit{Alternative monotone mappings}} \\

$\{1,2,3\}$
& $552.0$
& $482.0$
& $454.5\ (+17.7\%)$
& $449.8\ (+18.5\%)$
& $398.2\ (+27.9\%)$
& $393.2\ (+28.8\%)$ \\

$\{1,2,5\}$
& $634.0$
& $551.5$
& $497.5\ (+21.5\%)$
& $486.2\ (+23.3\%)$
& $448.2\ (+29.3\%)$
& $435.2\ (+31.3\%)$ \\

$\{1,3,10\}$
& $1025.2$
& $888.8$
& $771.8\ (+24.7\%)$
& $746.8\ (+27.2\%)$
& $697.0\ (+32.0\%)$
& $672.8\ (+34.4\%)$ \\

$\{0,1,5\}$
& $391.2$
& $337.2$
& $274.2\ (+29.9\%)$
& $260.5\ (+33.4\%)$
& $249.5\ (+36.2\%)$
& $236.5\ (+39.6\%)$ \\

\bottomrule
\end{tabular}
\end{adjustbox}

\end{table}
\paragraph{Continuous severity weights.}
We also elicit continuous $0$--$100$ severity scores from two judge
families, \texttt{deepseek-chat} and GPT-4o, on both SWE pools. The
continuous scores correlate with the ordinal labels
($\rho=0.38$--$0.49$ across judge-pool combinations) and with each
other (cross-judge Spearman $0.50$ on Lite and $0.40$ on Verified).
Re-running allocation with continuous scores as loss weights preserves
the ordering in all four combinations (Table~\ref{tab:contcost}).

\begin{table}[h]
\centering
\caption{\textbf{Allocation under continuous $0$--$100$ severity weights.}
Values report loss reduction relative to difficulty-aware routing.
The deployment route remains the ordinal issue-only predictor, while
loss weights and oracle routing use continuous severity.}
\label{tab:contcost}

\small
\renewcommand{\arraystretch}{1.10}
\setlength{\tabcolsep}{4.5pt}

\begin{adjustbox}{width=\linewidth}
\begin{tabular}{@{}l@{\hspace{7pt}}r@{\hspace{7pt}}rr@{\hspace{7pt}}rr@{}}
\toprule
&
\multicolumn{1}{c}{\textbf{Baseline}}
&
\multicolumn{2}{c}{\textbf{Consequence-aware}}
&
\multicolumn{2}{c}{\textbf{Priority-aware}} \\
\cmidrule(lr){2-2}
\cmidrule(lr){3-4}
\cmidrule(l){5-6}

Severity weights
& Random
& Pred.
& Cont.\ oracle
& Pred.
& Cont.\ oracle \\
\midrule

\rowcolor{gray!12}
\multicolumn{6}{c}{\textit{SWE-bench Lite}} \\

deepseek-chat
& $+13.1\%$
& $\mathbf{+16.7\%}$
& $+20.6\%$
& $+25.3\%$
& $+31.5\%$ \\

GPT-4o
& $+13.9\%$
& $\mathbf{+17.8\%}$
& $+20.8\%$
& $+27.6\%$
& $+30.7\%$ \\

\midrule
\rowcolor{gray!12}
\multicolumn{6}{c}{\textit{SWE-bench Verified}} \\

deepseek-chat
& $+12.5\%$
& $\mathbf{+14.2\%}$
& $+18.7\%$
& $+26.4\%$
& $+30.9\%$ \\

GPT-4o
& $+13.7\%$
& $\mathbf{+13.8\%}$
& $+22.6\%$
& $+25.2\%$
& $+30.1\%$ \\

\bottomrule
\end{tabular}
\end{adjustbox}

\end{table}
\section{Predictor Calibration Details}
\label{app:predictor}

This appendix provides full confusion matrices,
per-class metrics, operating thresholds, and cross-pool safety
statistics for the deployment-time consequence predictor.
\begin{figure}[h]
\centering
\includegraphics[width=0.95\linewidth]{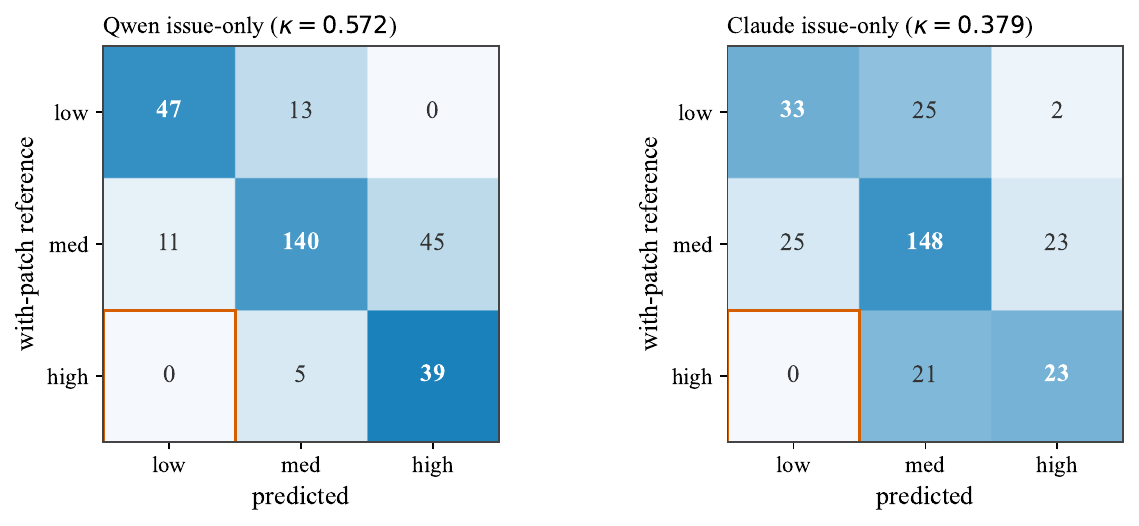}
\caption{\textbf{Predictor confusion matrices} against the
LLM-with-patch reference on 300 SWE-bench Lite tasks. Rows denote the
reference class and columns denote the predicted class. The
deployment-critical under-allocation cell, true class~2 and predicted
class~0, is empty for both predictors.}
\label{fig:confusion}
\end{figure}
\paragraph{Qwen issue-only predictor.}
Table~\ref{tab:qwen-predictor} reports the full confusion matrix and
per-class performance of the Qwen issue-only predictor.

\begin{table}[h]
\centering
\caption{\textbf{Qwen issue-only predictor on SWE-bench Lite.}
Against the Qwen2.5-7B-Instruct with-patch reference on $300$ tasks,
the issue-only predictor achieves Cohen's $\kappa=0.572$.}
\label{tab:qwen-predictor}

\small
\renewcommand{\arraystretch}{1.10}

\begin{tabular*}{\linewidth}{@{\extracolsep{\fill}}lrrrr@{}}
\toprule

\rowcolor{gray!12}
\multicolumn{5}{c}{\textit{Confusion matrix}} \\

&
\multicolumn{3}{c}{\textbf{Predicted label}}
& \\
\cmidrule(lr){2-4}

Reference label
& $\hat{s}=0$
& $\hat{s}=1$
& $\hat{s}=2$
& Total \\
\midrule

$s=0$
& $47$
& $13$
& $0$
& $60$ \\

$s=1$
& $11$
& $140$
& $45$
& $196$ \\

$s=2$
& $\mathbf{0}$
& $5$
& $39$
& $44$ \\

\midrule

Total
& $58$
& $158$
& $84$
& $300$ \\

\midrule
\rowcolor{gray!12}
\multicolumn{5}{c}{\textit{Per-class performance}} \\

Class
& Precision
& Recall
& F1
& Support \\
\midrule

Class 0
& $0.81$
& $0.78$
& $0.80$
& $60$ \\

Class 1
& $0.89$
& $0.71$
& $0.79$
& $196$ \\

Class 2
& $0.46$
& $\mathbf{0.89}$
& $0.61$
& $44$ \\

\midrule
\rowcolor{gray!12}
\multicolumn{5}{c}{\textit{Safety summary}} \\

High-to-low errors ($2\!\to\!0$)
& \multicolumn{4}{c}{$\mathbf{0/44}$} \\

\bottomrule
\end{tabular*}

\end{table}
\paragraph{Claude issue-only predictor.}
Table~\ref{tab:claude-predictor} reports the full confusion matrix and
per-class performance of the cross-model Claude Sonnet~4.5 predictor.

\begin{table}[h]
\centering
\caption{\textbf{Claude issue-only predictor on SWE-bench Lite.}
The cross-model Claude Sonnet~4.5 predictor achieves Cohen's
$\kappa=0.379$ against the with-patch reference. Notably, no
reference class-2 task is predicted as class~0 ($2\!\to\!0=0/44$).}
\label{tab:claude-predictor}

\small
\renewcommand{\arraystretch}{1.10}

\begin{tabular*}{\linewidth}{@{\extracolsep{\fill}}lrrrr@{}}
\toprule

\rowcolor{gray!12}
\multicolumn{5}{c}{\textit{Confusion matrix}} \\

&
\multicolumn{3}{c}{\textbf{Predicted label}}
& \\
\cmidrule(lr){2-4}

Reference label
& $\hat{s}=0$
& $\hat{s}=1$
& $\hat{s}=2$
& Total \\
\midrule

$s=0$
& $33$
& $25$
& $2$
& $60$ \\

$s=1$
& $25$
& $148$
& $23$
& $196$ \\

$s=2$
& $\mathbf{0}$
& $21$
& $23$
& $44$ \\

\midrule
Total
& $58$
& $194$
& $48$
& $300$ \\

\midrule
\rowcolor{gray!12}
\multicolumn{5}{c}{\textit{Per-class performance}} \\

Class
& Precision
& Recall
& F1
& Support \\
\midrule

Class 0
& $0.57$
& $0.55$
& $0.56$
& $60$ \\

Class 1
& $0.76$
& $0.76$
& $0.76$
& $196$ \\

Class 2
& $0.48$
& $0.52$
& $0.50$
& $44$ \\

\bottomrule
\end{tabular*}

\end{table}


\paragraph{Operating thresholds across pools.}
Table~\ref{tab:thresholds} compares two operating thresholds of the
ordinal consequence predictor for detecting class-$2$ tasks.

\begin{table}[h]
\centering
\caption{\textbf{Operating thresholds across evaluation pools.}
Recall and precision for detecting reference class-$2$ tasks when the
ordinal predictor is thresholded at $\hat{s}=2$ or $\hat{s}\geq1$.}
\label{tab:thresholds}

\small
\renewcommand{\arraystretch}{1.10}
\setlength{\tabcolsep}{6pt}

\begin{adjustbox}{width=\linewidth}
\begin{tabular}{@{}lrcc@{}}
\toprule
&
&
\multicolumn{2}{c}{\textbf{Detection threshold}} \\
\cmidrule(lr){3-4}

Pool
& True class-2
& $\hat{s}=2$
& $\hat{s}\geq1$ \\
&
&
Recall / Precision
& Recall / Precision \\
\midrule

\rowcolor{gray!12}
\multicolumn{4}{c}{\textit{Software-engineering pools}} \\

SWE-bench Lite
& $44$
& $0.89$ / $0.46$
& $\mathbf{1.00}$ / $0.18$ \\

MSWE-bench mini
& $3$
& $1.00$ / $0.16$
& $\mathbf{1.00}$ / $0.01$ \\

SWE-bench Verified
& $47$
& $0.96$ / $0.62$
& $\mathbf{1.00}$ / $0.37$ \\

\midrule
\rowcolor{gray!12}
\multicolumn{4}{c}{\textit{Out-of-domain pools}} \\

BIRD
& $3$
& $1.00$ / $0.20$
& $\mathbf{1.00}$ / $0.03$ \\

FinQA
& $10$
& $0.60$ / $1.00$
& $\mathbf{1.00}$ / $0.09$ \\

MATH
& $3$
& $0.00$ / ---
& $\mathbf{1.00}$ / $0.04$ \\

\midrule
\rowcolor{gray!12}
\multicolumn{4}{c}{\textit{Cross-pool safety summary}} \\

All six pools
& $110$
& \multicolumn{2}{c}{
$2\!\to\!0=\mathbf{0/110}$;
one-sided $95\%$ upper bound $=2.7\%$
} \\

\bottomrule
\end{tabular}
\end{adjustbox}

\end{table}
\section{Per-Task Marginal Utility and Marginal-Gain Prediction}
\label{app:marginal}

This appendix supports the method and discussion around the exact
priority rule
\[
\mathrm{priority}(x)=w_x\Delta_x.
\]
It documents per-task marginal utility, deployment-time attempts to
predict marginal gain, and the role of execution feedback.

\paragraph{Per-task examples.}
Table~\ref{tab:marg} contrasts tasks selected by priority with tasks
selected by raw difficulty.

\begin{table}[h]
\centering
\caption{\textbf{Per-task marginal utility excerpt.}
Priority-ranked tasks combine substantial consequence with high marginal
compute gain, whereas difficulty-ranked tasks lie in the unsalvageable tail
where additional compute provides no observed gain.}
\label{tab:marg}

\small
\renewcommand{\arraystretch}{1.10}

\begin{tabular*}{\linewidth}{@{\extracolsep{\fill}}lcccc@{}}
\toprule
Selection
& Consequence
& Difficulty
& $\Delta_x$
& Priority score \\
\midrule

\rowcolor{gray!12}
\multicolumn{5}{c}{\textit{Top-8 by priority}} \\

top-1
& $2$
& $0.50$
& $\mathbf{1.00}$
& $\mathbf{2.00}$ \\

top-2
& $2$
& $0.50$
& $\mathbf{1.00}$
& $\mathbf{2.00}$ \\

top-3
& $2$
& $0.56$
& $0.75$
& $1.50$ \\

top-4
& $2$
& $0.56$
& $0.75$
& $1.50$ \\

top-5
& $2$
& $0.62$
& $0.75$
& $1.50$ \\

top-6
& $2$
& $0.62$
& $0.75$
& $1.50$ \\

top-7
& $1$
& $0.31$
& $\mathbf{1.00}$
& $1.00$ \\

top-8
& $1$
& $0.31$
& $\mathbf{1.00}$
& $1.00$ \\

\midrule
\rowcolor{gray!12}
\multicolumn{5}{c}{\textit{Top-4 by difficulty}} \\

top-1
& $1$
& $1.00$
& $\mathbf{0.00}$
& $\mathbf{0.00}$ \\

top-2
& $1$
& $1.00$
& $\mathbf{0.00}$
& $\mathbf{0.00}$ \\

top-3
& $2$
& $1.00$
& $\mathbf{0.00}$
& $\mathbf{0.00}$ \\

top-4
& $0$
& $1.00$
& $\mathbf{0.00}$
& $\mathbf{0.00}$ \\

\bottomrule
\end{tabular*}

\end{table}
\paragraph{Prompted marginal-gain prediction.}
We prompted \texttt{deepseek-chat} and GPT-4o to forecast the success
probability of a basic agent and a frontier agent from issue text and
file path alone. The absolute success levels are moderately
predictable, with $\rho$ up to $+0.39$, but the difference
$\hat{\Delta}_x$ is not: correlations with true
$\Delta_x$ lie between $-0.07$ and $+0.05$ across
family-pool combinations.

\paragraph{Learned marginal-gain probe.}
A ridge regression probe on issue-text embeddings
(\texttt{text-embedding-3-small}, 10-fold task-level cross-validation)
also fails to predict marginal gain reliably:
$\rho(\hat{\Delta}_x,\Delta_x)=+0.11$ on Lite ($p=0.06$) and
$+0.13$ on Verified (n.s.). AUC for detecting $\Delta_x>0$ is $0.49$
on Lite and $0.34$ on Verified, and cross-pool transfer is near zero.

\paragraph{Execution feedback carries signal.}
After a first failed attempt, a gold-free verifier score becomes
informative. Among tasks whose first attempt fails, the verifier score
predicts whether attempts $2$--$8$ rescue the task with
$\rho=+0.23$ ($p=0.016$, AUC $0.65$) under the Claude judge and
$\rho=+0.21$ ($p=0.027$, AUC $0.65$) under the DeepSeek judge. This
supports the distinction used in the main text: marginal gain is hard
to estimate before execution, but in-flight feedback can support
sequential allocation.

\section{Why Difficulty-Aware Allocation Fails}
\label{app:whyfail}

This appendix provides the full marginal-gain analysis summarized in
\S\ref{sec:main}.

\paragraph{Marginal gain.}
Let
\[
\Delta_x
=
p_{\mathrm{premium}}(x)-p_{\mathrm{cheap}}(x).
\]
This quantity measures where premium compute changes the outcome.

\begin{figure}[h]
\centering
\includegraphics[width=\linewidth]{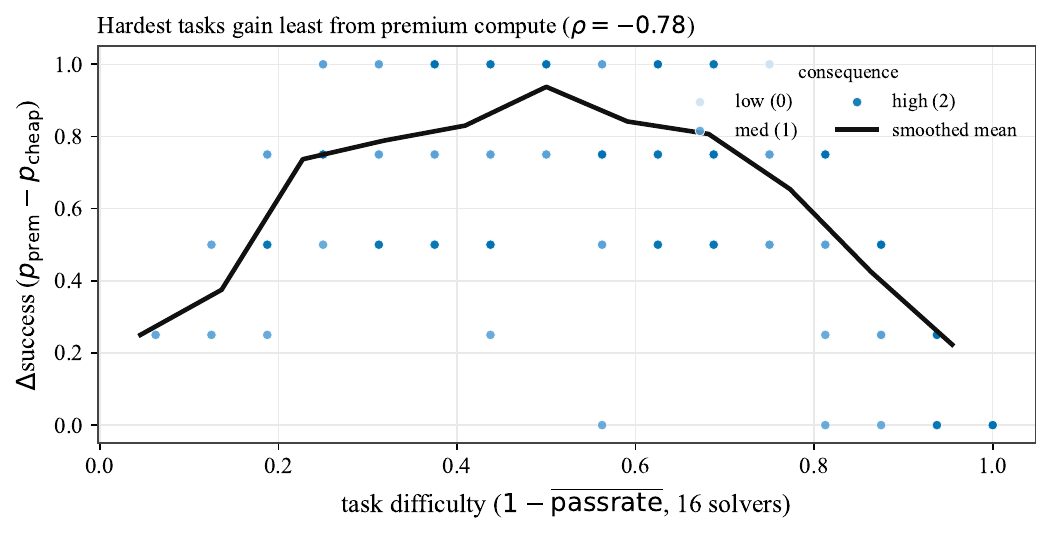}
\caption{\textbf{Difficulty and marginal gain on SWE-bench Lite.}
Marginal gain collapses on the hardest tasks, producing
$\rho(\mathrm{difficulty},\Delta_x)=-0.78$
($p<10^{-60}$).}
\label{fig:marginal-app}
\end{figure}

\paragraph{Difficulty anti-selects marginal gain.}
On SWE-bench Lite,
\[
\rho(\mathrm{difficulty},\Delta_x)=-0.78,
\qquad p<10^{-60}.
\]
Many of the hardest tasks remain unsolved even at the premium tier, so
upgrading them consumes compute without changing outcomes.

\paragraph{Not only a never-solved artifact.}
There are $76/300$ tasks solved by no system. Removing them, the
correlation remains strongly negative:
\[
\rho=-0.489,\qquad p<10^{-14}.
\]
Thus, the pattern is not purely caused by the never-solved floor.

\paragraph{Deployment-time difficulty estimates.}
Issue-only solvability forecasts from two model families produce
difficulty estimates whose correlation with marginal gain is weak and
not significant: $-0.087$ and $+0.058$ on Lite, and $-0.051$ and
$-0.144$ on Verified. Routing by these deployment-time difficulty
estimates lands near random and below consequence routing.

\paragraph{Consequence and marginal gain.}
Consequence is nearly uncorrelated with marginal gain:
\[
\rho(\mathrm{consequence},\Delta_x)=+0.09
\quad \text{(n.s.)}.
\]
Consequence therefore does not predict where compute helps. It predicts
where help is valuable if it occurs.

\paragraph{Consequence is not reducible to confidence.}
We also test whether consequence is merely a proxy for cross-model
disagreement. Let
\[
Y(x)=w_x\Delta_x
\]
be the cost-weighted marginal value of upgrading task $x$. We regress
$Y$ on difficulty, pass-rate variance, and consequence. Since $Y$
contains $w_x$ by construction, this analysis should be read only as
a non-collinearity check, not as evidence that consequence predicts
marginal gain. The result shows that consequence is not absorbed by
difficulty or pass-rate variance (Table~\ref{tab:r2decomp}).
\begin{table}[h]
\centering
\caption{\textbf{$R^2$ decomposition of cost-weighted marginal value.}
We regress $Y=w_x\Delta_x$ on difficulty, pass-rate variance, and
consequence. Because $Y$ contains consequence multiplicatively, the
increment from adding consequence is partly mechanical; this analysis is
used only as a non-collinearity check.}
\label{tab:r2decomp}

\small
\renewcommand{\arraystretch}{1.12}

\begin{tabular*}{\linewidth}{@{\extracolsep{\fill}}lc@{}}
\toprule
Predictors
& $R^2$ \\
\midrule

\rowcolor{gray!12}
\multicolumn{2}{c}{\textit{Single predictors}} \\

Difficulty
& $0.281$ \\

Pass-rate variance
& $0.474$ \\

Consequence
& $0.350$ \\

\midrule
\rowcolor{gray!12}
\multicolumn{2}{c}{\textit{Combined predictors}} \\

Difficulty + pass-rate variance
& $0.474$ \\

Difficulty + pass-rate variance + consequence
& $\mathbf{0.738}$ \\

\midrule
\rowcolor{gray!12}
\multicolumn{2}{c}{\textit{Incremental contribution}} \\

$\Delta R^2$ from adding consequence
& $\mathbf{+0.264}$ \\

\bottomrule
\end{tabular*}

\end{table}
\paragraph{Boundary conditions.}
The same decomposition predicts when consequence routing should fail:
when costs are homogeneous, when premium compute has no headroom, or
when difficulty is positively coupled with marginal gain.
Appendix~\ref{app:ood} tests these cases out of domain.

\section{Robustness Across SWE-bench Sub-Domains}
\label{app:subdomain}

This appendix checks whether the main findings are driven by a
particular sub-domain of SWE-bench Lite. We define heuristic
sub-domains using lexical filters and rerun orthogonality, predictor
agreement, and allocation analyses inside each group.
\begin{table}[h]
\centering
\caption{\textbf{Robustness across SWE-bench Lite sub-domains.}
Within each sub-domain, $\rho$ measures consequence--difficulty
correlation, $\kappa$ measures issue-only predictor agreement with the
with-patch reference, $2\!\to\!0$ counts high-to-low prediction errors,
and $\Delta$ vs.\ Diff. reports the priority-aware oracle gain over
difficulty-aware routing.}
\label{tab:subdomain}

\small
\renewcommand{\arraystretch}{1.10}
\setlength{\tabcolsep}{4.5pt}

\begin{adjustbox}{width=\linewidth}
\begin{tabular}{@{}lrrrrr@{}}
\toprule
&
&
\multicolumn{1}{c}{\textbf{Orthogonality}}
&
\multicolumn{1}{c}{\textbf{Prediction}}
&
\multicolumn{1}{c}{\textbf{Safety}}
&
\multicolumn{1}{c}{\textbf{Allocation}} \\

Sub-domain
& $n$
& $\rho(\mathrm{cons.},\mathrm{diff.})$
& $\kappa$
& $2\!\to\!0$
& $\Delta$ vs.\ Diff. \\
\midrule

Numeric / scientific computing
& $36$
& $-0.13$
& $0.52$
& $\mathbf{0}$
& $+30.2\%$ \\

Parsers / CLI / tokens
& $35$
& $-0.10$
& $0.69$
& $\mathbf{0}$
& $+33.1\%$ \\

Rendering / UI / formatting
& $61$
& $+0.13$
& $0.61$
& $\mathbf{0}$
& $+30.6\%$ \\

Imports / modules / configuration
& $141$
& $-0.08$
& $0.53$
& $\mathbf{0}$
& $+30.7\%$ \\

\bottomrule
\end{tabular}
\end{adjustbox}

\end{table}
Table~\ref{tab:subdomain} shows that the same pattern holds across
sub-domains: consequence remains weakly related to difficulty, the
predictor avoids high-to-low errors, and priority-aware routing improves
over difficulty-aware routing.

\section{Cross-Dataset Transfer: Multi-SWE-bench mini}
\label{app:crossdataset}

This appendix evaluates transfer of the consequence construct and
issue-only predictor to Multi-SWE-bench mini.

\paragraph{Pool structure.}
The pool is structurally low-consequence. The with-patch judge assigns
only $3/400$ tasks ($0.75\%$) to class~2, compared with $44/300$
($14.7\%$) on SWE-bench Lite. Most tasks are class~1.

\paragraph{Predictor transfer.}
Table~\ref{tab:transfer-mswe} reports the full cross-pool transfer
results of the Qwen issue-only predictor.

\begin{table}[h]
\centering
\caption{\textbf{Qwen issue-only predictor transfer.}
On $399$ doubly labeled tasks, the predictor achieves Cohen's
$\kappa=0.382$ with $72.2\%$ raw agreement. All three reference
class-$2$ tasks are recovered, with no high-to-low errors.}
\label{tab:transfer-mswe}

\small
\renewcommand{\arraystretch}{1.10}

\begin{tabular*}{\linewidth}{@{\extracolsep{\fill}}lrrrr@{}}
\toprule

\rowcolor{gray!12}
\multicolumn{5}{c}{\textit{Confusion matrix}} \\

&
\multicolumn{3}{c}{\textbf{Predicted label}}
& \\
\cmidrule(lr){2-4}

Reference label
& $\hat{s}=0$
& $\hat{s}=1$
& $\hat{s}=2$
& Total \\
\midrule

$s=0$
& $67$
& $39$
& $1$
& $107$ \\

$s=1$
& $56$
& $218$
& $15$
& $289$ \\

$s=2$
& $\mathbf{0}$
& $0$
& $\mathbf{3}$
& $3$ \\

\midrule
Total
& $123$
& $257$
& $19$
& $399$ \\

\midrule
\rowcolor{gray!12}
\multicolumn{5}{c}{\textit{Transfer summary}} \\

Cohen's $\kappa$
& \multicolumn{4}{c}{$0.382$} \\

Raw agreement
& \multicolumn{4}{c}{$72.2\%$} \\

Class-2 recall
& \multicolumn{4}{c}{$\mathbf{100\%}$ $(3/3)$} \\

High-to-low errors ($2\!\to\!0$)
& \multicolumn{4}{c}{$\mathbf{0/3}$} \\

\bottomrule
\end{tabular*}

\end{table}

\paragraph{Why no allocation study.}
We do not run a matched-compute allocation study on Multi-SWE-bench
mini because no public per-task multi-model outcome table is available
for constructing cheap and premium tiers, and because the pool contains
only three high-consequence tasks. Appendix~\ref{app:seconddomain}
therefore provides the second full software allocation study on a
class-$2$-rich database/ORM pool.

\section{A Second Software Pool: Database/ORM Tasks from SWE-bench Verified}
\label{app:seconddomain}

This appendix repeats the full allocation comparison on a disjoint
pool of $140$ database/ORM tasks from SWE-bench Verified.

\paragraph{Pool construction.}
We select \texttt{django/django} tasks whose issue text or touched
files match database/ORM signals: migrations, schema, SQL, querysets,
transactions, integrity constraints, and related terms. We exclude all
instances overlapping the main SWE-bench Lite pool. Compute tiers are
built from $20$ public SWE-bench Verified leaderboard submissions,
using the bottom four systems as cheap tier and the top four as
premium tier.

\paragraph{Consequence distribution.}
The with-patch judge assigns $47/140$ tasks ($33.6\%$) to class~2,
making this a consequence-rich software pool. The full distribution is
$\{0{:}8,\ 1{:}85,\ 2{:}47\}$.

\paragraph{Predictor transfer.}
Table~\ref{tab:transfer-verified} reports the transfer performance of
the Qwen issue-only predictor on the SWE-bench Verified database/ORM pool.

\begin{table}[h]
\centering
\caption{\textbf{Qwen issue-only predictor transfer on SWE-bench Verified.}
The predictor achieves Cohen's $\kappa=0.582$ with $75.7\%$ raw
agreement. It recovers $45/47$ reference class-$2$ tasks and makes no
high-to-low errors.}
\label{tab:transfer-verified}

\small
\renewcommand{\arraystretch}{1.10}

\begin{tabular*}{\linewidth}{@{\extracolsep{\fill}}lrrrr@{}}
\toprule

\rowcolor{gray!12}
\multicolumn{5}{c}{\textit{Confusion matrix}} \\

&
\multicolumn{3}{c}{\textbf{Predicted label}}
& \\
\cmidrule(lr){2-4}

Reference label
& $\hat{s}=0$
& $\hat{s}=1$
& $\hat{s}=2$
& Total \\
\midrule

$s=0$
& $8$
& $0$
& $0$
& $8$ \\

$s=1$
& $4$
& $53$
& $28$
& $85$ \\

$s=2$
& $\mathbf{0}$
& $2$
& $\mathbf{45}$
& $47$ \\

\midrule

Total
& $12$
& $55$
& $73$
& $140$ \\

\midrule
\rowcolor{gray!12}
\multicolumn{5}{c}{\textit{Transfer summary}} \\

Cohen's $\kappa$
& \multicolumn{4}{c}{$0.582$} \\

Raw agreement
& \multicolumn{4}{c}{$75.7\%$} \\

Class-2 recall
& \multicolumn{4}{c}{$\mathbf{95.7\%}$ $(45/47)$} \\

High-to-low errors ($2\!\to\!0$)
& \multicolumn{4}{c}{$\mathbf{0/47}$} \\

\bottomrule
\end{tabular*}

\end{table}
\paragraph{Allocation replication.}
Table~\ref{tab:seconddomain} repeats the matched-compute comparison.

\begin{table}[h]
\centering
\caption{\textbf{Second-pool allocation on SWE-bench Verified.}
Matched-compute allocation results on $140$ database/ORM tasks using
an independent $20$-system compute tier.}
\label{tab:seconddomain}

\small
\renewcommand{\arraystretch}{1.10}

\begin{tabular*}{\linewidth}{@{\extracolsep{\fill}}lrr@{}}
\toprule
Strategy
& $\mathcal{L}$
& $\Delta$ vs.\ Diff. \\
\midrule

\rowcolor{gray!12}
\multicolumn{3}{c}{\textit{Difficulty-based and random baselines}} \\

Difficulty-aware (oracle difficulty)
& $154.75$
& $0\%$ \\

Random (mean over $1{,}000$ draws)
& $135.61$
& $+12.4\%$ \\

\midrule
\rowcolor{gray!12}
\multicolumn{3}{c}{\textit{Consequence-aware routing}} \\

\textbf{Consequence-aware (Qwen issue-only)}
& $131.75$
& $\mathbf{+14.9\%}$ \\

Consequence-aware (with-patch oracle)
& $121.00$
& $+21.8\%$ \\

\midrule
\rowcolor{gray!12}
\multicolumn{3}{c}{\textit{Priority-aware analysis}} \\

Priority-aware (pred.\ $\times$ empirical gain)
& $111.25$
& $+28.1\%$ \\

Priority-aware (oracle $\times$ empirical gain)
& $\mathbf{106.00}$
& $\mathbf{+31.5\%}$ \\

\bottomrule
\end{tabular*}

\end{table}

The strategy ordering exactly replicates the main pool:
difficulty-aware is worst, random is better, consequence-aware improves
further, and priority-aware is best.

\section{Judge Robustness for Consequence Labels}
\label{app:judge2}

This appendix tests whether the results depend on the Qwen2.5-7B
with-patch judge. We relabel both SWE pools with
\texttt{deepseek-chat} using the same with-patch rubric prompt.

\paragraph{Judges calibrate severity differently.}
On SWE-bench Lite, Qwen and DeepSeek agree at $\kappa=0.182$ with raw
agreement $48.0\%$. DeepSeek is much more severe, assigning $54.7\%$
of Lite tasks to class~2 versus $14.7\%$ for Qwen. On the Verified
database/ORM pool, agreement is $\kappa=0.140$.

\paragraph{Ordering is mostly preserved.}
On Lite under DeepSeek weights, difficulty-aware remains worst
($435.25$), random improves ($378.31$, $+13.1\%$), consequence-aware
improves further ($362.75$, $+16.7\%$), and priority-aware is best
(up to $305.75$, $+29.8\%$). On Verified, difficulty-aware is also
worst and priority-aware best, while consequence-only matches random
because DeepSeek assigns class~2 to most of the pool, flattening cost
heterogeneity.

\paragraph{Why the consequence-only margin can collapse.}
This collapse is predicted by the objective. Consequence routing can
beat random only by concentrating premium compute on tasks that carry
disproportionate cost weight. When a judge marks most tasks as high
consequence, the top-quartile weight share approaches the uniform
baseline of $0.25$, leaving little room for a consequence ranking to
outperform random. In our judge-swap analysis, the top-quartile weight
share tracks the consequence-over-random margin: settings with strong
cost heterogeneity produce a larger consequence advantage, while the
DeepSeek Verified setting flattens the weights and therefore collapses
the consequence-only margin.

\paragraph{Interpretation.}
Judge swaps reveal a threshold-calibration issue, not a collapse of
the construct. When a judge flattens the cost distribution by marking
most tasks high consequence, consequence-only routing has little room
to outperform random. This is consistent with the objective: the value
of consequence routing depends on perceived cost heterogeneity.

\section{Out-of-Domain Probes: BIRD, FinQA, and MATH}
\label{app:ood}

Figure~\ref{fig:generalization}(a,c) summarizes the main
out-of-domain boundary-condition results. This appendix provides the
full dataset-specific protocols, allocation tables, and declared-stakes
sweeps.

\paragraph{BIRD text-to-SQL.}
We sample $150$ questions from BIRD dev, stratified over business
databases and official difficulty labels. The cheap tier is GPT-4o-mini
and the premium tier is o4-mini, with three samples per task per tier.
All generations are graded by official execution accuracy. Execution
accuracy rises from $44.0\%$ to $53.3\%$, giving modest headroom. The
question-only consequence predictor agrees with the with-gold judge at
$\kappa=0.568$ and recovers all three class-$2$ tasks under the primary
Qwen labeling.
Under the primary Qwen weights, costs are nearly homogeneous:
only $3/150$ tasks are class~2. Consequence routing therefore collapses
toward random, as predicted. Under a GPT-4o judge that assigns higher
stakes to medical, toxicological, and financial queries, consequence
becomes more useful.

\begin{table}[h]
\centering
\caption{\textbf{BIRD text-to-SQL allocation under alternative consequence weights.}
Cost-weighted loss is evaluated under Qwen- and GPT-4o-derived consequence
weights. $\Delta$ reports the relative reduction in loss versus
difficulty-aware routing under the same weighting.}
\label{tab:ood-bird}

\small
\renewcommand{\arraystretch}{1.10}
\setlength{\tabcolsep}{4.5pt}

\begin{adjustbox}{width=\linewidth}
\begin{tabular}{@{}lrr@{\hspace{7pt}}rr@{}}
\toprule
&
\multicolumn{2}{c}{\textbf{Qwen weights}}
&
\multicolumn{2}{c}{\textbf{GPT-4o weights}} \\
\cmidrule(lr){2-3}
\cmidrule(l){4-5}

Strategy
& $\mathcal{L}$
& $\Delta$ vs.\ Diff.
& $\mathcal{L}$
& $\Delta$ vs.\ Diff. \\
\midrule

\rowcolor{gray!12}
\multicolumn{5}{c}{\textit{Difficulty-based and random baselines}} \\

Difficulty-aware
& $66.67$
& $0\%$
& $75.00$
& $0\%$ \\

Random
& $68.27$
& $-2.4\%$
& $75.80$
& $-1.1\%$ \\

\midrule
\rowcolor{gray!12}
\multicolumn{5}{c}{\textit{Consequence-aware routing}} \\

Consequence-aware (question-only pred.)
& $67.33$
& $-1.0\%$
& $72.67$
& $+3.1\%$ \\

Consequence-aware (with-gold oracle)
& $65.67$
& $+1.5\%$
& $66.67$
& $+11.1\%$ \\

\midrule
\rowcolor{gray!12}
\multicolumn{5}{c}{\textit{Priority-aware analysis}} \\

Priority-aware (pred.\ $\times$ gain)
& $58.00$
& $+13.0\%$
& $\mathbf{62.00}$
& $\mathbf{+17.3\%}$ \\

Priority-aware (oracle $\times$ gain)
& $\mathbf{56.33}$
& $\mathbf{+15.5\%}$
& $\mathbf{62.00}$
& $\mathbf{+17.3\%}$ \\

\bottomrule
\end{tabular}
\end{adjustbox}

\end{table}

\paragraph{FinQA.}
On $150$ FinQA questions, generations are graded by numeric match with
standard percent/decimal normalization. Premium reasoning compute
provides almost no headroom: o4-mini scores below the cheap tier
($64.2\%$ vs $66.2\%$), and GPT-4o adds only $+1.1$ points. The
issue-only predictor reaches $\kappa=0.661$ against the with-context
judge, but with no meaningful premium headroom all routing signals land
near random. This is exactly the objective's prediction: if upgrading
rarely helps, choosing where to upgrade cannot matter.

\paragraph{MATH.}
On $150$ MATH problems, generations are graded by normalized
boxed-answer match. The predictor reaches $\kappa=0.514$ against the
with-problem judge. Premium compute has large headroom
($65.3\%\to82.3\%$), and difficulty is positively coupled with
marginal gain:
\[
\rho(\mathrm{level},\Delta_x)=+0.257,
\qquad p=0.002.
\]
Here difficulty routing wins, because harder math problems are exactly
where the premium reasoning tier helps. This is the opposite regime
from SWE-bench.

\paragraph{Boundary-condition summary.}

\begin{table}[h]
\centering
\caption{\textbf{Boundary-condition summary.}}
\label{tab:boundary-summary}

\small
\renewcommand{\arraystretch}{1.10}
\setlength{\tabcolsep}{4.5pt}

\begin{adjustbox}{width=\linewidth}
\begin{tabular}{@{}lcccl@{}}
\toprule
Pool
& Top-$25\%$ weight share
& $\rho(\mathrm{diff.},\Delta_x)$
& Headroom
& Best deployable signal \\
\midrule

SWE Lite
& $0.42$
& $-0.78$
& Yes
& Consequence \\

SWE Verified DB/ORM
& $0.39$
& $-0.38$
& Yes
& Consequence \\

BIRD
& $0.35$
& $+0.06$
& Modest
& None / near random \\

FinQA
& $0.36$
& $+0.01$
& None
& None \\

MATH
& $0.50$
& $+0.26$
& Large
& Difficulty \\

\bottomrule
\end{tabular}
\end{adjustbox}

\end{table}

\paragraph{Declared dollar stakes.}
We additionally run a controlled stakes sweep on FinQA and MATH. Stakes
tiers are assigned orthogonally to difficulty, and the cost spread
between low and high stakes is swept from $1\times$ to $10^6\times$.

\begin{table}[h]
\centering
\caption{\textbf{Declared-stakes sensitivity.}
Stakes are assigned orthogonally to difficulty while the high-to-low
cost spread $S$ increases from $1\times$ to $10^{6}\times$.
Positive margins indicate lower loss for consequence-aware than
difficulty-aware routing.}
\label{tab:stakescross}

\small
\renewcommand{\arraystretch}{1.10}
\setlength{\tabcolsep}{4.5pt}

\begin{adjustbox}{width=\linewidth}
\begin{tabular}{@{}lrrrrrrr@{}}
\toprule
Cost spread $S$
& $1\times$
& $3\times$
& $10\times$
& $30\times$
& $100\times$
& $10^{3}\times$
& $10^{6}\times$ \\
\midrule

\rowcolor{gray!12}
\multicolumn{8}{c}{\textit{MATH}} \\

Margin vs.\ difficulty
& $-28.3\%$
& $-18.4\%$
& $-6.5\%$
& $\mathbf{+3.2\%}$
& $+11.5\%$
& $+20.2\%$
& $+24.9\%$ \\

Cons.\ wins
& $0\%$
& $0\%$
& $26\%$
& $\mathbf{62\%}$
& $76\%$
& $84\%$
& $84\%$ \\

\midrule
\rowcolor{gray!12}
\multicolumn{8}{c}{\textit{FinQA}} \\

Margin vs.\ difficulty
& $+2.9\%$
& $+3.6\%$
& $+4.4\%$
& $+5.1\%$
& $+5.7\%$
& $+6.3\%$
& $+6.6\%$ \\

Cons.\ wins
& $84\%$
& $78\%$
& $76\%$
& $76\%$
& $76\%$
& $76\%$
& $76\%$ \\

\bottomrule
\end{tabular}
\end{adjustbox}

\end{table}

The sweep quantifies when cost heterogeneity can override difficulty:
on MATH, consequence routing flips from losing to winning between
$10\times$ and $30\times$ cost spread.

\section{Within-Model Allocation: Attempts as the Compute Lever}
\label{app:within}

This appendix provides the full protocol for the controlled
within-model experiment summarized in \S\ref{sec:main}.

\paragraph{Thinking-token budget as a weak lever.}
We first test thinking-token budgets on two families. For Claude
Sonnet~4.5, sweeping budgets
$\{1024,2048,4096,8192,16000\}$ on an $80$-task stratified sample
produces no consistent success improvement; pass rates are
$0.100/0.064/0.100/0.150/0.100$. For DeepSeek-R1, increasing hard caps
$\{4096,8192,16000,24000\}$ raises success only from $1/80$ to
$7/80$, with zero class-$2$ successes at every cap. Budget-level
allocation is therefore degenerate on these tasks.

\paragraph{Attempts as a real compute lever.}
We fix a Claude Sonnet~4.5 agent with the same prompt and thinking
budget and vary only the number of attempts. On the scaled sample,
best-of-$k$ pass rate rises from $\mathrm{pass@}1=0.142$ to
$\mathrm{pass@}8=0.370$, giving real compute headroom.

\paragraph{Original 48-task experiment.}
In the original sample, $48$ tasks receive all eight attempts. The
matched budget is $216$ attempts: $24$ tasks at best-of-$8$ and
$24$ tasks at a single attempt. Results are:

\begin{table}[h]
\centering
\caption{\textbf{Original within-model allocation on 48 tasks.}
All strategies use the same $216$-attempt budget, with $24$ tasks
allocated best-of-$8$ and the remaining $24$ tasks a single attempt.}
\label{tab:within}

\small
\renewcommand{\arraystretch}{1.10}

\begin{tabular*}{\linewidth}{@{\extracolsep{\fill}}lrr@{}}
\toprule
Strategy
& $\mathcal{L}$
& $\Delta$ vs.\ Diff. \\
\midrule

\rowcolor{gray!12}
\multicolumn{3}{c}{\textit{Baselines}} \\

Difficulty-aware
& $36.0$
& $0\%$ \\

Random
& $30.0$
& $+16.7\%$ \\

\midrule
\rowcolor{gray!12}
\multicolumn{3}{c}{\textit{Consequence-aware allocation}} \\

\textbf{Consequence-aware (issue-only predictor)}
& $29.0$
& $\mathbf{+19.4\%}$ \\

Consequence-aware (oracle)
& $27.0$
& $+25.0\%$ \\

\midrule
\rowcolor{gray!12}
\multicolumn{3}{c}{\textit{Priority-aware analysis}} \\

Priority-aware (pred.\ $\times$ gain)
& $27.0$
& $+25.0\%$ \\

Priority-aware (oracle)
& $\mathbf{26.0}$
& $\mathbf{+27.8\%}$ \\

\bottomrule
\end{tabular*}

\end{table}

Bootstrap intervals are wide but positive: $[+6.5,+44.1]\%$ for
consequence-aware and $[+11.6,+47.2]\%$ for priority-aware.

\paragraph{Scaled 127-task experiment.}
We extend to $127$ tasks with all eight attempts under the Claude
judge. Difficulty-aware loss is $102.00$, random $91.92$,
consequence-aware issue-only $83.00$ ($+18.6\%$), oracle consequence
$85.00$ ($+16.7\%$), and priority-aware predictor $77.00$
($+24.5\%$). The task-level bootstrap interval is
$[+10.5,+30.7]\%$ for consequence-aware and $[+14.6,+35.6]\%$ for
priority-aware. A stricter DeepSeek judge on $126$ tasks preserves the
ordering with smaller margins: consequence-predictor $+8.7\%$ and
priority $+13.5\%$.

\paragraph{Dynamic caps with oracle verifier.}
We also evaluate sequential policies offline. Attempts are issued one
at a time and stop at first success. Consequence-gated caps dominate:
under the Claude judge, a deployment-time profile
$k_{\max}=2/8/8$ by predicted class matches uniform best-of-$8$ loss
while consuming $20\%$ fewer attempts; the oracle-gated version saves
$28\%$. Difficulty-gated caps do not match uniform-$8$ loss at lower
budget.

\paragraph{Gold-free verifier.}
Replacing the oracle verifier with a gold-free \texttt{deepseek-chat}
verifier, consequence-gated caps match uniform best-of-$8$ loss at
$31$--$42\%$ less realized compute across judges and acceptance
thresholds. Difficulty-gating adds little beyond a uniform lower cap.
This is the fully deployable sequential stack: issue-only consequence
predictor, gold-free verifier, and consequence-gated attempt caps.
Because the gold-free verifier is also a \texttt{deepseek-chat} model,
the DeepSeek-judge arm may benefit from same-family error correlation;
the Claude-judge arm is the cleaner cross-family evaluation and
preserves the same conclusion.

\paragraph{Judge agreement.}
On the original attempt set, Claude and DeepSeek judges agree on
$91.3\%$ of attempts with Cohen's $\kappa=0.630$. Recomputing the
allocation comparison under DeepSeek verdicts preserves the same
ordering.

\begin{figure}[h]
\centering
\includegraphics[width=\linewidth]{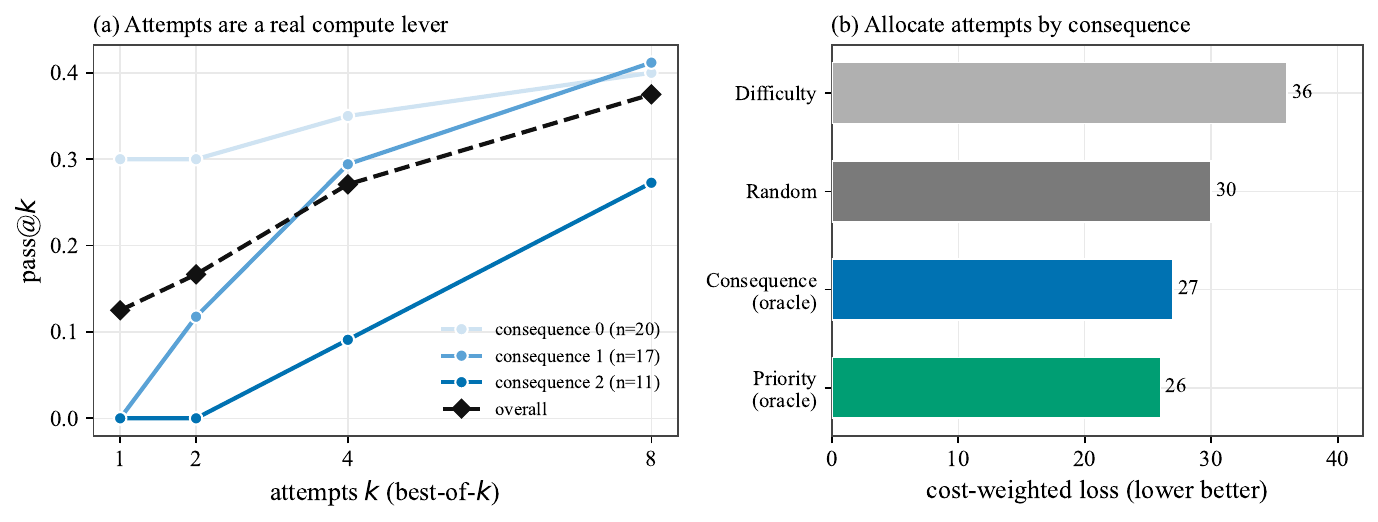}
\caption{\textbf{Within-model allocation with a single fixed agent.}
Left: best-of-$k$ pass rate rises with the number of attempts, giving
real within-model compute headroom. Right: at a matched total attempt
budget, consequence-aware and priority-aware allocation reduce
cost-weighted loss below difficulty-aware routing.}
\label{fig:within}
\end{figure}

\section{A Live Online Deployment Run}
\label{app:online}

This appendix evaluates whether the routing machinery can run as a
live streaming system rather than an offline replay.

\paragraph{Setup.}
A stream of $300$ tasks, $150$ FinQA plus $150$ MATH, arrives one at a
time. Tasks carry deployment-declared stakes mapped to
\$10K/\$1M/\$10M. Four policies process the same stream under a
real-time premium budget: random, difficulty, consequence, and a
verifier-gated sequential policy. The run uses live calls to
\texttt{gpt-4o-mini}, \texttt{gpt-4o}, and \texttt{o4-mini}. Routing
components never see gold answers; grading is post-hoc.

\paragraph{Results.}
\begin{table}[h]
\centering
\caption{\textbf{Live online deployment run.}
Results on a $300$-task FinQA+MATH stream under real-time premium-budget
pacing. Loss reduction is measured relative to random routing.}
\label{tab:online}

\small
\renewcommand{\arraystretch}{1.10}

\begin{tabular*}{\linewidth}{@{\extracolsep{\fill}}lrrrr@{}}
\toprule
&
\multicolumn{2}{c}{\textbf{Loss}}
&
\multicolumn{2}{c}{\textbf{Online operation}} \\
\cmidrule(lr){2-3}
\cmidrule(l){4-5}

Policy
& DWL
& Reduction vs.\ random
& Premium calls
& Accuracy \\
\midrule

\textsc{Random}
& $24{,}426$
& $0\%$
& $23.0\%$
& $70.0\%$ \\

\textsc{Diff}
& $\mathbf{16{,}432}$
& $\mathbf{+32.7\%}$
& $25.3\%$
& $70.7\%$ \\

\textsc{Cons}
& $19{,}831$
& $+18.8\%$
& $19.3\%$
& $68.7\%$ \\

\textsc{Seq}
& $21{,}627$
& $+11.5\%$
& $\mathbf{5.0\%}$
& $70.0\%$ \\

\bottomrule
\end{tabular*}

\end{table}

The stream lies in a regime where the framework predicts difficulty,
not consequence, should dominate because MATH has positive
difficulty--gain coupling. The live run confirms this. The
verifier-gated sequential policy uses only $5.0\%$ premium calls while
matching random accuracy, showing that the machinery can operate online
with real-time budget pacing.

\paragraph{Scope.}
This is not a live consequence-wins demonstration. It shows that the
routing system works online and respects the predicted boundary
conditions. A production online run in the SWE-style regime remains
future work.

\section{Context-Conditional Consequence}
\label{app:context}

Consequence depends on deployment context. The same error can be low
severity in a hobby project and high severity in a payment or
safety-critical system. This appendix treats context as an explicit
input:
\[
s_x \rightarrow s_{x,\mathrm{context}}.
\]

\paragraph{Setup.}
For $100$ stratified SWE-bench Lite tasks, we elicit continuous
severity scores under four declared deployment contexts: a personal
hobby project, an internal business tool with human review, a
payment-processing service, and a safety-critical industrial control
system. We use two judge families, \texttt{deepseek-chat} and GPT-4o,
in both with-patch and issue-only modes, producing $1600$ labels.

\paragraph{Severity shifts monotonically with context.}
Mean severity rises monotonically along the stakes ladder. For
DeepSeek with-patch labels, the means are
$22.9 \to 26.2 \to 50.6 \to 64.7$. For GPT-4o, they are
$27.4 \to 28.4 \to 34.2 \to 42.5$. The shift is often monotone within
individual tasks as well: non-decreasing severity appears for
$71\%$ of tasks under DeepSeek and $90\%$ under GPT-4o in judge mode.

\paragraph{Routing adapts to context.}
Using each context's with-patch severity as the weight and the same
context's issue-only severity as the routing signal, consequence
routing beats random in all eight context-by-judge cells
($+6.1\%$ to $+10.9\%$) and difficulty routing in all eight
($+17.5\%$ to $+23.5\%$). The selected premium sets differ across
contexts: overlap between the top-$25\%$ sets under hobby and
safety-critical contexts is only $36\%$ under DeepSeek and $52\%$
under GPT-4o.

\paragraph{Interpretation.}
Context-dependence is not a threat to consequence-aware allocation; it
is an input to it. The scheduler can condition on deployment context
when the deployer provides it.

\section{Deployment-Time Consequence-Only Scheduler}
\label{app:scheduler}

This appendix provides the implementation details of the
deployment-time consequence-only scheduler introduced in
\S\ref{sec:method} and used in our experiments.
The scheduler is intentionally lightweight: it does not modify the
underlying reasoning model, require additional training, or use
execution feedback. Instead, it only determines how an existing
compute budget is distributed across tasks based on predicted
consequence.
Given a set of tasks $X$, a deployment-time consequence predictor
produces $\hat{s}_x$, which is mapped to
$\hat{w}_x=g(\hat{s}_x)$. Tasks are then ranked by $\hat{w}_x$, and higher-ranked tasks are upgraded
to stronger compute tiers whenever additional budget is available.
The scheduler starts from the cheapest tier and greedily allocates
remaining budget to the tasks with the highest predicted consequence.
The procedure is designed to match the deployment setting considered
in this paper. At routing time, the scheduler only has access to
information available before solving, including the issue description
and pre-solution metadata used by the consequence predictor. It does
not access gold patches, generated solutions, execution outcomes, or
any post-hoc information. Therefore, the algorithm represents a pure
ex-ante compute allocation strategy.

\begin{algorithm}[t]
\caption{Deployment-time consequence-only compute allocation}
\label{alg:cons-route}
\begin{algorithmic}[1]
\REQUIRE Tasks $X$; deployment-time consequence predictor $\hat{s}$;
tiers $T_1,\ldots,T_K$ with costs $\kappa_k$; total budget $B$.

\STATE Compute predicted weights:
$\hat{w}_x\leftarrow g(\hat{s}_x)$ for all $x\in X$

\STATE Sort tasks in descending order of $\hat{w}_x$

\STATE Initialize all tasks with the cheapest tier:
$T(x)\leftarrow T_1$

\STATE Initialize current cost:
$C\leftarrow |X|\kappa_1$

\FOR{$k=K$ down to $2$}
    \FOR{$x$ in sorted task order}
        \IF{$T(x)=T_1$ and
        $C+\kappa_k-\kappa_1\le B$}
            \STATE Upgrade task $x$ to tier $T_k$
            \STATE Update budget:
            $C\leftarrow C+\kappa_k-\kappa_1$
        \ENDIF
    \ENDFOR
\ENDFOR

\RETURN Allocation $T$
\end{algorithmic}
\end{algorithm}

The greedy procedure in Algorithm~\ref{alg:cons-route}  is not intended as an optimal solver for all
possible budget allocation problems. Its purpose is to provide a
simple and deployable scheduler that isolates the contribution of
consequence as a routing signal. More sophisticated optimization
methods can be incorporated when additional information, such as
execution feedback or marginal compute gains, becomes available.

\end{document}

%% file: math_commands.tex
\usepackage{amsmath,amsfonts,bm}

\def\eqref#1{equation~\ref{#1}}

\def\1{\bm{1}}

\DeclareMathAlphabet{\mathsfit}{\encodingdefault}{\sfdefault}{m}{sl}
\SetMathAlphabet{\mathsfit}{bold}{\encodingdefault}{\sfdefault}{bx}{n}

